\documentclass{article}

\PassOptionsToPackage{numbers, compress}{natbib}
\usepackage[preprint]{neurips_2020}

\usepackage[utf8]{inputenc} % allow utf-8 input
\usepackage[T1]{fontenc}    % use 8-bit T1 fonts
\usepackage{hyperref}       % hyperlinks
\usepackage{url}            % simple URL typesetting
\usepackage{booktabs}       % professional-quality tables
\usepackage{amsfonts}       % blackboard math symbols
\usepackage{nicefrac}       % compact symbols for 1/2, etc.
\usepackage{microtype}      % microtypography
\usepackage{graphicx}  
\usepackage{caption}
\usepackage{multirow}

\title{On the Demystification of Knowledge Distillation: A Residual Network Perspective}

\author{
  Nandan Kumar Jha\thanks {equal contribution}\\
  IIT Hyderabad\\
  Telangana, India 502285 \\
  \texttt{cs17mtech11010@iith.ac.in} \\ 
  \And
  Rajat Saini$^*$ \\
  TRI-AD \\
  Tokyo, Japan 103-0022 \\
  \texttt{rajat.saini@tri-ad.global} \\
  \And
  Sparsh Mittal \\
  IIT Roorkee \\
  Uttarakhand, India 247667\\
  \texttt{sparsh.mittal@ece.iitr.ac.in} \\
}

\begin{document}

\maketitle

\begin{abstract}

Knowledge distillation (KD) is generally considered as a technique for performing model compression and learned-label smoothing. However, in this paper, we study and investigate the KD approach from a new perspective: we study its efficacy in training a deeper network without any residual connections. We find that in most of the cases, non-residual student networks perform equally or better than their residual versions trained on raw data without KD (baseline network). {\em Surprisingly}, in some cases, they surpass the accuracy of baseline networks even with the inferior teachers. After a certain depth of non-residual student network, the accuracy drop, coming from the removal of residual connections, is substantial, and training with KD boosts the accuracy of the student up to a great extent; however, it does not fully recover the accuracy drop. Furthermore, we observe that the conventional teacher-student view of KD is incomplete and does not adequately explain our findings. We propose a novel interpretation of KD with the \emph{Trainee-Mentor} hypothesis, which provides a holistic view of KD. We also present two viewpoints, loss landscape, and feature reuse, to explain the interplay between residual connections and KD. We substantiate our claims through extensive experiments on residual networks.

\end{abstract}

\section{Introduction}
Knowledge distillation (KD) \cite{hinton2015distilling} has been proposed to transfer the representations learned by a large and higher-capacity (teacher) network to a  shallower (student) network. The success of KD as a model compression technique is believed to be in ``dark knowledge'' encoded in teacher's logits ({\em soft targets}). From the network's training standpoint, KD serves as a learned level smoothing approach \cite{tang2020understanding,yuan2020revisiting}  and enables faster convergence \cite{phuong2019towards,yim2017gift}.  However, Furlanello et al. \cite{furlanello2018born} showed that an identically parametrized student performs much better than the teacher network. 
Moreover, even when the teacher has lower predictive performance or is poorly trained, the use of KD increases the student's performance more than when it is trained only on raw data (hard targets) \cite{yuan2020revisiting}. 
In light of these findings, we claim that the ``student-teacher'' learning paradigm is an incomplete view of KD.
Furthermore, previous research on KD does not adequately explain the desirable properties of students and teachers that can find the best student-teacher pairs. This limitation is especially troublesome for the case of inferior teacher \cite{furlanello2018born,yuan2020revisiting} where there is a minimal understanding about ``{\em when and why the students may become better than the teacher?}''. To summarize, there is a need for a holistic understanding of KD.

It is well-known that training a deeper network is challenging for many reasons:   deterioration of propagating signal quality diminishes feature reuse in the forward pass, vanishing/exploding gradients in the backward pass, and increased non-linearity \cite{li2018visualizing}. Residual connections  resolve the issue of vanishing/exploding gradients \cite{srivastava2015highway,he2016deep,he2016identity} and prevent explosion of non-convexity which increases the non-chaotic regions on error surface \cite{li2018visualizing}. Apart from residual connections, a better initialization also helps to stabilize the training of deeper networks \cite{glorot2010understanding,mishkin2015all,zhang2019fixup}. From a loss landscape standpoint, a better initialization leads to convergence in the convex portion of the error surface where minima with low loss value are present \cite{li2018visualizing}.

\textbf{Contributions:} In this work, we propose a ``{\em Trainee-Mentor}'' (TM) hypothesis, which leads to a novel interpretation of KD and offers a holistic understanding of the efficacy of KD. We argue that in KD, the teacher does not share his knowledge, which is implicit as structured representation. Instead, the teacher offers a hint and guides the training of the student network. This hint leads to a {\em better weight initialization} in the student network during the initial phase of training, and the network starts converging in a well-behaved region of loss-landscape, i.e., in a potentially good basin of attraction. To validate our hypothesis, we revisit the KD and disentangle its applicability as a model compression technique or a learned/adaptive regularizer. We see KD as a substitute for residual connections in training a deeper network. 

We take a residual network as the baseline. We obtain its non-residual version by removing all the residual connections, which leads to accuracy loss. This network acts as a student in our study. The teacher is a variant of the baseline residual network. Through our systematic experimental study, we make the following observations. (1) shallower non-residual networks do not merely recover the accuracy drop incurred due to the removal of residual connections, but in some cases, they surpass the accuracy of both their baseline residual network and the teacher network; (2) when the depth of non-residual student network increases beyond a certain limit, its accuracy increases to a great extent, however, use of KD cannot fully recover the accuracy loss from the removal of residual connections. These findings naturally lead to the following questions: {\em what purpose does the KD serve in effectively training the shallower plain (non-residual) networks? Why does it fail in a deeper network? What are the desirable properties of a student-teacher pair such that the student can perform even better than the teacher?}. 
 
We seek to answer these questions through our proposed TM hypothesis, where we liken the representational power of the network to ``implicit knowledge'' and regularization to the ``experience''. 
We hypothesize that implicit knowledge in a network is {\em quite structural} since it is gained through a step-wise learning process during the training phase. By contrast, learning through regularization enhances the network's experience by breaking the co-adaption between neurons, thus improving the generalization power. Based on the insights gained from TM hypothesis, we claim the following: (1) In KD, teacher (mentor) does not share/transfer his knowledge to a student (trainee) instead it gives a hint, which leads to a better initialization in the very beginning phase of training. This allows the student network to start converging in a well-defined non-chaotic region on the error surface; (2) A student who uses experience (regularized network) can get more benefit from teacher's hint unless the teacher network itself is overly-regularized; (3) when the teacher is overly-regularized, a student with higher knowledge (i.e., higher representational power) gets more benefit from the teacher's hint. Further, we show that the KD is a better substitute for residual connections in a shallower network from the loss-landscape and feature reuse viewpoint. Our contribution can be summarized as follows. 

\begin{enumerate}
\item We propose ``{\em Trainee-Mentor} hypothesis, which provides a holistic understanding of KD and presents general principles for KD's effectiveness. This hypothesis helps explain the case of inferior teachers, where the student gets a boost in accuracy even with the inferior teacher.
\item We present a novel interpretation of KD as a substitute for residual connection in shallower residual connections.
\item We have done extensive experiments with Wide ResNet (WRN) \cite{zagoruyko2016wide}, ResNet \cite{he2016deep}, and MobileNetV2 \cite{sandler2018mobilenetv2} on CIFAR-100 \cite{2012KrizhevskyCIFAR} dataset to validate our hypothesis and claims.
\end{enumerate}

\section{Related Work}

{\bf Importance of residual connections and their substitutes} 
As proposed in \cite{he2016deep}, residual connection solves the degradation problem (in networks with increasing depth, error rate gets first saturated and then increased) by eliminating the optimization difficulty, which causes slower convergence in deeper networks, and facilitates training of deeper networks. 
\cite{balduzzi2017shattered} shows that residual connections cause sub-linear decay in gradient, in contrast to exponential decay in non-residual feed-forward networks, and prevents gradient shattering in deeper networks. \cite{philipp2018gradients} claim that identity skip connection solves gradient exploding problem and enables effective training of deeper networks. \cite{orhan2017skip} and \cite{li2018visualizing} demonstrated the efficacy of residual connections from {\em loss landscape} viewpoint. \cite{orhan2017skip} argued that residual connection eliminates the singularities, which causes learning slow-down, inherent in the loss landscape of deeper networks.  \cite{li2018visualizing} showed that residual connections smoothen the loss landscape by eliminating the chaotic regions on the error surface. \cite{li2017convergence} studied the importance of identity mapping and on a two-layer ResNet with ReLU activation and showed that the presence of identity mapping removes spurious local minima and saddle points on error surface. Also, \cite{liu2019towards} demonstrated that random initialized feed-forward CNNs could be trapped in the spurious local optimum with a probability of at least $\frac{1}{4}$. Whereas,  residual network with appropriate step size and initialization starts sufficiently away from the spurious local optimum and enters in the basin of attraction of global optimum and converge to it. \cite{veit2016residual} introduced ``unraveled view'' which suggests that a residual network can be viewed as an ensemble of exponential number ($O(2^n)$)  of shallower networks with different path-lengths, where path-length follows a binomial distribution, instead of an ultra-deep network. They argued that residual connection does not solve the gradient vanishing problem in deeper networks; rather, it avoids the gradient vanishing problem as most of the gradient comes from relatively shorter paths.

On the {\em representational} effect of residual connections, \cite{greff2016highway} presented an ``unrolled iterative estimation'' viewpoint of learning representation in residual blocks and showed that the layers in residual blocks iteratively refine their estimates of the same features instead of computing an entirely new representation. \cite{philipp2018gradients} and \cite{zagoruyko2016wide} argue that gradient stability comes at the cost of loss in representational power in deeper residual networks.  On that note, \cite{huang2016deep} proposed ``stochastic depth'' where they train a deeper residual network with randomly dropped layers in each epoch and achieve a significant improvement in error rate. Also, \cite{zhang2019revisiting} proposed a nonlinear ReLU group normalization shortcut, which alleviates the representational power degradation problem of identity skip connection by effectively utilizing the representation of shallower networks and perform better than deeper networks with identity skip connections.

Deeper networks can also be trained without any shortcut connections. In congruence with the unraveled view presented in \cite{veit2016residual}, \cite{larsson2016fractalnet} proposed an alternative to residual network: ``Fractal architecture'' based FractalNet which contains subnetworks with different path length (depth). \cite{zagoruyko2017diracnets} proposed ``Dirac weight parameterization'' enabling the training of deeper networks without the need for explicit residual connections and also eliminates the burden of careful initialization in both residual and non-residual networks. \cite{monti2018avoiding} showed that skip connection plays an important role in the initial phase of training, and they gradually removed the skip-connections in the later phase of training.

{\bf Knowledge distillation}
There has been a great body of work for understanding the efficacy of KD. \cite{vapnik2015learning} argued that intelligent teachers provide privileged information, only available at the training phase, to the students. Later, \cite{lopez2015unifying} proposed generalized distillation, where KD is being unified with privileged information. \cite{gotmare2018a} demonstrated that KD has a similar effect as learning rate warmup.  They empirically showed that learning rate warmup prevents gradient instability caused by deeper layers; similarly, the effect of dark knowledge localized majorly in deeper (discriminative) layers, in contrast to the initial (feature extraction) layers.  \cite{furlanello2018born} studied the gradient flow in KD and compared with that of in normal training and found that KD acts as importance weighting, to each sample in a mini-batch, where weights represent the teacher's confidence in the correct predictions. \cite{zhang2018deep} presented the posterior entropy viewpoint of KD and claimed that ``soft-target'' enables robustness by regularizing the much more informed choice of alternatives than blind entropy regularization. A recent work \cite{phuong2019towards} attributed the success of KD to the optimization bias and showed that gradient descent has a favorable bias for distillation, and a favorable data geometry (angular alignment between data distribution and the teacher).

\cite{cho2019efficacy} and \cite{mirzadeh2019improved} showed the ineffectiveness of KD when the gap between student and teacher capacity is high. To overcome this challenge, the former used the early-stopping (in training with KD) while the latter introduced intermediate teachers as teaching assistants and bridge the capacity gap between student and teacher.  \cite{yuan2020revisiting} claimed the success of KD is not only due to the similarity information of output categories but also due to the inherent regularization (learned level smoothing regularization) enabled by ``soft-targets'' from teachers. They showed that students could benefit from weak and poorly trained teachers whose representational power is inferior to that of the students. A recent work \cite{cheng2020explaining} hypothesized that, unlike normal training, network in KD learns only task-relevant visual concepts, simultaneously in each epoch of training,  and also learn more foreground visual concepts. There is also a great body of work on understanding the efficacy of self-distillation (when both students and teachers have identical architecture). For example, \cite{mobahi2020self} revealed that self-distillation acts as a regularizer that progressively limits the number of basis functions (in Hilbert space) that can represent the solution.

There is another line of research to improve upon the conventional KD. \cite{romero2014fitnets} proposed FitNets and showed that in addition to the output logits, intermediate feature representations from teacher model significantly improve the performance of thin and deep student networks (FitNets). \cite{ahn2019variational} proposed variational information distillation where the mutual information between students and teachers are maximized.  This outperforms the traditional KD, especially for the distillation across the heterogeneous architectures (e.g., CNN and MLP). \cite{yang2019training} analyzed KD from  ``strictness'' of the teacher's perspective and claimed that less strict teacher has less peaked distributions of confidence and preserves secondary (image-dependent) information, enables the student to learn inter-class similarity. This makes students stronger by preventing being fit to unnecessary strict distribution. A recent work \cite{liu2020search} claimed that knowledge of a (teacher) network depends on both its parameters as well as its architecture, and hence best student architecture for different teachers could differ. They proposed architecture-aware knowledge distillation, using neural architecture search, to find the best student model for a given teacher. \cite{passalis2020heterogeneous} claimed that traditional KD is unaware of information plasticity (temporal distribution of information in the networks \cite{achille2018critical}), and they proposed critical learning-period aware KD.

\section{Methodology} 

\subsection{Architecture of Networks: Representational power and Generalization Ability} \label{subsec:ArchitectureOfNetwork}

\begin{figure}[htbp]
\centering
\includegraphics[scale=0.7]{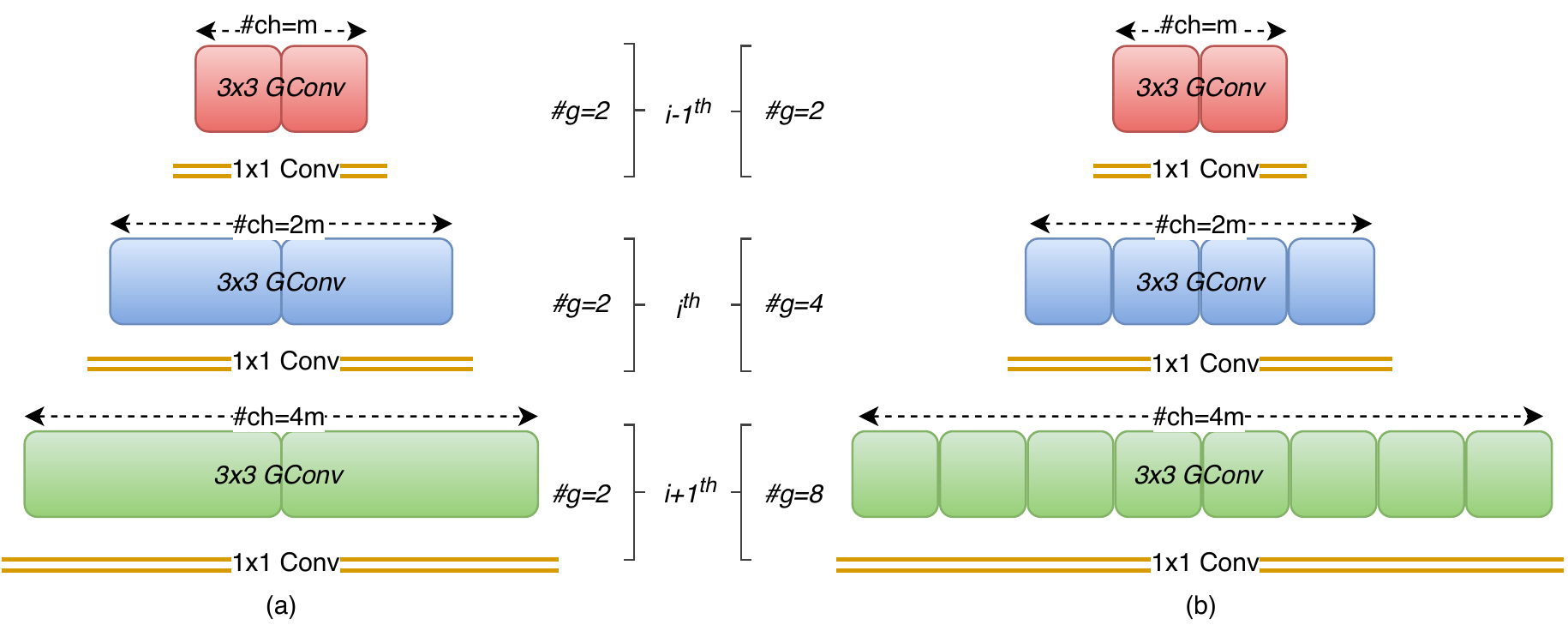} 
\caption{ Group and Groupsize illustration}
\label{fig:GroupIllustration}
\end{figure}

Let $n$, $m$, $k\times k$, $f\times f$ be the number of output feature maps (ofmap), the number of input feature map (ifmap), spatial size of the kernel, and spatial size feature map (fmap) are respectively. We employ group convolution (GConv) \cite{krizhevsky2012imagenet,xie2017aggregated,zhang2018shufflenet} with two variants constant $g$, and constant $G$ where the former is the number of groups, and the latter is the number of channels per groups in a convolutional layer (Fig. \ref{fig:GroupIllustration}). The number of channels in each group of a network with constant $g$ is $G= \frac{m}{g}$. The number of parameters for standard convolution and GConv are $n\times m\times k^2$ and $\frac{n\times m\times k^2}{g}$ respectively. That is, increasing $g$ or decreasing $G$ reduces the number of parameters. Depth-wise convolution ($G$=1) \cite{howard2017mobilenets} has lowest number of parameters, whereas, standard convolution ($g$=1) has the highest number of parameters.  
In all the tables, student/teacher with ``R'' and ``NR'' prefix represents the residual and non-residual (respectively) version of the original network. Accuracy drop is calculated over the difference between the accuracy of baseline (student) residual network and their non-residual version trained on ``hard-target''.
Distillation gain is computed as the gain over baseline residual network trained on ``hard-targets'' when the non-residual version of the student is trained in KD with the various residual network as a teacher.  

We have selected group convolution for our experiments due to the following reasons: (1) changing the number of groups ($g$ or $G$) alters the representational power of the network. For example, more number of channels in the group of deeper convolutional layers capture a more latent concept of a complex object and improve the performance \cite{ioannou2017deep}; (2) lower $G$ or higher $g$ results into fewer channels per group and break the filter co-dependence and improves the generalization \cite{ioannou2017deep}. In summary, just by tweaking $g$ or $G$, one can get numerous of a network with similar architecture and enables a trade-off between representational power and generalization ability.  To validate our hypothesis and claims, we need to have neural networks with different degrees of regularization (experience) and representational power (knowledge). Additionally, GConv is proved to be amenable for KD \cite{crowley2018moonshine}.

\subsection{Trainee-Mentor Hypothesis} \label{subsec:TMHypothesis}

In this section, first, we present the interpretation of training neural networks from the viewpoint of getting knowledge and experience to solve a complex task. Then, through the same viewpoint, we explain the fallacy of the Teacher-Student view of KD. Further, we present a holistic viewpoint of KD with our proposed ``{\em Trainee-Mentor}'' (TM) hypothesis.  

{\bf Knowledge and experience learned by a neural network} Training a neural network is analogous to learning to solve a complex task where input is a problem statement (question), parametric layers represent the intermediate steps for solving the problem, and output is a solution (answer). The correctness of the solution is assessed through the ``empirical risk minimization'' of loss (usually the cross-entropy loss) function, calculated over one-hot encoded hard target. For this purpose, some variants of a first-order optimization technique such as stochastic gradient descent (SGD) are used. Based on the feedback propagated in the network through back-propagation, the network updates its weights and refines the internal representations, which is equivalent to refining the intermediate steps for solving the given problem. Thus, the network enhances its representational power (knowledge base for solving a problem) in each iteration of training. In a sense, the implicit knowledge in a network is equivalent to its representational power, and hence, a more abstract representation implies detailed knowledge. In conclusion, {\em the inherent knowledge in a trained network is quite structural}. 

Notice that the learning capacity of a network greatly depends on its architecture. For example, due to higher capacity in deeper networks, it learns the representation with a higher number of abstractions and achieves higher representational power. However, a wider network can cause over-fitting, which is equivalent to memorizing the answer to a question rather than learning a generalized and systematic approach to answer a question. To avoid this problem, the regularization methods such as dropout \cite{srivastava2014dropout}, drop-connect \cite{wan2013regularization}, drop-block \cite{ghiasi2018dropblock}, drop-path \cite{larsson2016fractalnet}, etc., are employed during training phase to break the co-adaption among the learning entities such as neurons, layers, blocks, etc. Here, enabling regularization in the learning phase is equivalent to improving the network's experience in solving a problem. Thus, {\em the implicit experience gained during the training is quite unstructured}. Therefore, in addition to a great knowledge base, an experienced network can better answer an unseen question and improve the generalization.

{\bf Challenging the ``teacher-student'' viewpoint:} 
The ``dark knowledge'' contains the similarity information between output classes. Hence, unlike the cross-entropy loss, the ``{\em soft-target}''  in KD objective captures the correlation between output logits. This correlation is propagated along with the error gradients during the training in KD. Hence, a weak student learns the representation from a large capacity teacher and improves the accuracy as compared to the case when a student is trained without KD's objective. On the contrary, a significantly weak teacher, which has much less representational power than the student,  enables the improvement in student's performance \cite{yuan2020revisiting}.  Also, a student with similar representational power performs much better than the teacher when trained using KD \cite{furlanello2018born}. Due to this, there is no consensus about representational power's role in ``dark knowledge'', i.e., in the success of KD. Recent works attributed the success of KD to ``training-example re-weighting'' \cite{furlanello2018born}, and  ``learned/adaptive level smoothing regularization'' \cite{yuan2020revisiting} property of dark knowledge instead of the representation transfer. 
The KD objective only captures the correlation between class logits \cite{tian2019contrastive}. It ignores the higher-order output dependencies since it considers all dimensions of the output as an independent. Therefore, a contrastive objective that captures both the logits correlation and higher-order output dependencies performs better than KD \cite{tian2019contrastive}. Also, a similar performance improvement over KD has been reported by previous works  \cite{yim2017gift,romero2014fitnets} when the distilled knowledge considers the representation from intermediate layers in addition to the class logits.

\begin{table} [htbp]
\caption{Accuracy drop and distillation gain in ResNet18: students NR-g8, and NR-g16 in left table, whereas, all students in right table not only recover the accuracy drop, {\em they achieve significant positive gain over  baseline models}.}
\label{tab:ResNet18}
\begin{minipage}{.47\linewidth}
%\centering 
\resizebox{1.0\textwidth}{!}{
\begin{tabular} {c|c|c|c|c} 
\toprule
$I_t$-$S_s$ & NR-g2 & NR-g4 & NR-g8 & NR-g16 \\ \toprule
R-G1 & 71.93 & 71.3 & 71.84 & 73.17 \\
R-G2 & 70.76 & 72.38 & 72.38 & 73.61 \\
R-G4 & 70.14 & 71.98 & 71.07 & 73.79 \\
R-G8 & 70.23 & 72.04 & 72.82 & 73.33 \\
R-G16 & 70.19 & 71.98 & 72.27 & 73.55 \\ \midrule
Baseline & 73.47 & 72.59 & 71.64 & 71.53 \\
NR-baseline & 66.14 & 66.43 & 69.24 & 70.47 \\ \midrule
Acc. drop ($\downarrow$)& 7.33 & 6.16 & 2.4 & 1.06 \\
Distil. gain ($\uparrow$)& -1.54 & -0.21 & {\bf 1.18} & {\bf 2.26} \\ \bottomrule
\end{tabular}}
\end{minipage}
\begin{minipage}{.53\linewidth}
%\centering
\resizebox{1.0\textwidth}{!}{
\begin{tabular} {c|c|c|c|c} 
\toprule
I.t-S.s & NR-G2 & NR-G4 & NR-G8 & NR-G16 \\ \toprule
R-g2 & 73.88 & 73.81 & 72.76 & 72.86 \\
R-g4 & 73.31 & 73.91 & 73.31 & 71.12 \\
R-g8 & 73.30 & 73.40 & 73.69 & 71.18 \\
R-g16 & 73.60 & 72.56 & 73.47 & 73.16 \\ \midrule
Baseline & 71.08 & 71.18 & 71.36 & 72.22 \\
NR-baseline & 69.72 & 69.45 & 69.22 & 68.58 \\ \midrule
Acc. drop ($\downarrow$)& 1.36 & 1.73 & 2.14 & 3.64 \\
Distil. gain ($\uparrow$) & {\bf 2.8} & {\bf 2.73} & {\bf 2.33} & {\bf 0.94} \\ \bottomrule
\end{tabular}}
\end{minipage}
\end{table}
\begin{table} [htbp]
\caption{Accuracy drop and distillation gain in MobileNetV2: students NR-g2 (left table) and NR-G16 (right table) {\em have maximum distillation gain}, but do not fully recover the accuracy drop.} 
\label{tab:MV2}
\begin{minipage}{.47\linewidth}
%\centering 
\resizebox{1.0\textwidth}{!}{
\begin{tabular} {c|c|c|c|c} 
\toprule
$I_t$-$S_s$ & NR-g2 & NR-g4 & NR-g8 & NR-g16 \\ \toprule
R-G1 & 67.3 & 67.32 & 67.09 & 67.85 \\
R-G2 & 67.38 & 66.86 & 65.51 & 67.27 \\
R-G4 & 67.54 & 67.28 & 67.4 & 67.55 \\
R-G8 & 67.44 & 66.5 & 67.02 & 67.4 \\
R-G16 & 67.67 & 67.15 & 67.65 & 65.93 \\ \midrule
Baseline & 68.11 & 69.52 & 69.93 & 70.25 \\
NR-baseline & 58.68 & 58.72 & 58.77 & 58.79 \\ \midrule
Acc. drop ($\downarrow$)& 9.43 & 10.8 & 11.16 & 11.46 \\
Distil. gain ($\uparrow$)& -0.44 & -2.2 & -2.28 & -2.4 \\ \bottomrule
\end{tabular}}
\end{minipage}
\begin{minipage}{.53\linewidth}
%\centering
\resizebox{1.0\textwidth}{!}{
\begin{tabular} {c|c|c|c|c} 
\toprule
I.t-S.s & NR-G2 & NR-G4 & NR-G8 & NR-G16 \\ \toprule
R-g2 & 66.21 & 65.88 & 66.01 & 67.66 \\
R-g4 & 67.07 & 67.37 & 67.40 & 67.55 \\
R-g8 & 67.54 & 67.63 & 67.80 & 67.58 \\
R-g16 & 67.62 & 67.45 & 68.13 & 68.33 \\ \midrule
Baseline & 70.36 & 70.28 & 70.25 & 68.88 \\
NR-baseline & 60.77 & 59.66 & 58.62 & 57.18 \\ \midrule
Acc. drop ($\downarrow$)& 9.59 & 10.62 & 11.63 & 11.7 \\
Distil. gain ($\uparrow$)& -2.74 & -2.65 & -2.12 & -0.55 \\  \bottomrule
\end{tabular}}
\end{minipage}
\end{table}

{\bf Importance of initialization and TM hypothesis}
During the learning process, the accuracy of an answer depends on both the knowledge and experience learned until that point of training.  However, the initialization plays a crucial role in the optimization process and convergence to a well-behaved minima. Similarly, when a trainee solves a problem, then initially, they have numerous approaches to start with. Once training selected an approach, it can only refine the steps involved in the taken approach. In other words, if the trainee had selected a wrong approach, he/she would never be able to answer the question with less error. However, a hint/guidance at the very early stage of learning would greatly help choose the correct approach, which can later be refined to find the correct answer. This resembles the optimization of a neural network from the viewpoint of traversing on the error surface. This error surface is divided into the regions with well-behaved minima and highly non-convex (chaotic) regions. Optimization begins in the well-behaved region with better initialization, and the network converges to minima, which generalizes very well. In summary, the hint given by the mentor would help in the network's initialization. Note that to maintain consistency and to avoid confusion, we do not replace the term student/teacher with a trainee/mentor in his paper.

From the general perception in a trainee-mentor relationship, a trainee with more experience (more regularized student network) often gets more benefit from the mentor's hint compared to other trainees having a similar level of knowledge. However, when a hint is highly specialized (overly-regularized teacher), a student with more knowledge gets more benefit from the hint in solving the question. These claims are confirmed from the results shown in  Table \ref{tab:ResNet18} and Table \ref{tab:MV2}.  In Table \ref{tab:ResNet18}, as the regularization in students (with increasing $g$ or decreasing $G$) increasing their distillation gain is also increasing. However, in Table \ref{tab:MV2}, distillation gain is maximum for the students with higher representation power, i.e., students with higher $G$ or lower $g$.  Notice that MobileNetV2 is an overly-regularized model as the accuracy of their baseline increases with an increase in regularization (i.e., with higher $g$ or lower $G$) (Table \ref{tab:MV2}). By contrast, the accuracy in Table \ref{tab:ResNet18} increases with the increase in representational power.

\section{Interplay Between Residual Connections and Knowledge Distillation} \label{sec:ResidualConnection}

{\bf Feature reuse viewpoint}
The residual connection in a residual network enables feature reuse in forward pass and facilitate gradient stabilization in the backward pass. As the depth of networks increases, both feature reuse and stability of gradients become crucial. It is well-known that residual connections in a residual network act as depth-regularizer, which creates an exponential number of shallower networks with varying depth and avoids the gradient vanishing/exploding problem in deep networks. The number of these shallower networks increases exponentially as the depth of residual networks increases. This not only enables shorter paths for gradient propagation; it also mitigates the issue of {\em diminishing feature reuse}, which is more pronounced in deeper networks. Knowledge distillation compensates for the gradient stabilization, enabled by residual connections in a residual network, in a non-residual student network by regularizing the learning-phase of student and penalizing the gradient flow. However, KD does not compensate for the feature reuse, which boosts the representational power of the network since KD's objective is unable to transfer the structured representation in a teacher to the student. As a result, when the network's depth increases beyond a certain limit, KD is unable to recover the accuracy drop incurs due to the removal of residual connections in the student network (Table \ref{tab:WRN40x2} and Table \ref{tab:MV2}).

{\bf Loss landscape viewpoint} 
The effect of depth has a dramatic effect on the geometry of loss landscape \cite{li2018visualizing} when shortcut connections are not used. Hence, due to the increasing non-linearity in deeper networks, the loss surface of networks transition from the well-behaved regions to highly chaotic regions. Gradient propagation through these chaotic regions does not lead to  (potential) global minima. Moreover, it can lead to an unstable training when gradients start moving in the direction where loss function becomes substantially large (or the minimizers which are quite sharp). Residual connections prevent the transition from a well-behaved region to a highly non-convex region when network depth is increased and prevent chaotic behavior. 

A better initialization can also resolve the chaotic behavior in a non-residual network up to a great extent. Effectively, good initialization facilitates the propagation of gradient (in the initial iterations) to a well-behaved region rather than non-convex regions. Hence, the network can easily converge to flat minima (with lower loss value) and generalize better. KD effectively regularizes the gradient flow right from the very beginning phase of student's learning. This serves as a better initialization and guides gradients in the right direction. However, when the network's (non-residual) depth increases beyond a certain limit, the loss-surface is dominated by a chaotic region, and even a proper initialization does not find the flatter minima. As a result, KD does not fully recover (through better initialization in student's network) the accuracy drop, due to the removal of residual connections, when students are deeper (Table \ref{tab:WRN40x2} and Table \ref{tab:MV2}).

\section{Experimental Results} \label{sec:ExperimentalResults}

{\bf Implementation details}  For all the experiments, including the training of baseline with/without residual connections (R/NR) on hard-targets and the KD experimentations,   we use SGD with a mini-batch of size 128, initial learning as $0.1$, momentum is fixed as $0.9$, and weight decay factor is used as $4e-4$. All the R/NR-baseline network is trained with SGD for $120$ epochs except for the distillation experiments of Mobilenet-V2, where students are trained for 300 epochs. The learning rate is reduced by a factor of $10$ after every $30^{th}$, $60^{th}$, and $90^{th}$ epochs for ResNet18, whereas for WRN it is reduced by a factor of $5$. For MobileNetV2, the learning rate is multiplied by a factor of $0.98$ after every epoch for better convergence. For knowledge distillation experiments, we set temperature value as $4$ and hyper-parameter alpha as $0.9$.  Experiments are performed with CIFAR-100 \cite{2012KrizhevskyCIFAR} image classification dataset, which has 100 output classes with per class 100 training images and 100 testing images. For data augmentation, we pad each input image with a $4\times4$ reflection of edge pixels. Further, we use a horizontal flip and then a random crop of $32\times32$. For experiments on WRN, a dropout of 0.3 has been used to regularize the network.

{\bf Network Descriptions}  
We use different versions of WRN \cite{zagoruyko2016wide}, and ResNet \cite{he2016deep} where we replace standard $3\times3$ convolution by $3\times3$ GConv (either with constant $g$ or $G$) followed by $1\times1$ convolution. In MobileNetV2 \cite{sandler2018mobilenetv2} we replace $3\times3$ depthwise convolution with $3\times3$ GConv with constant $g$/$G$. (for more details please see the Appendix).

\begin{itemize}
\item {\bf WRN}:  In case of WRN, depth of network is controlled by the number of convolutional blocks as $r = \frac{d-4}{6}$  where $n$ is the number of convolutional blocks and $d$ is the depth of the network \cite{zagoruyko2016wide}. As mentioned above, all the standard $3\times3$ convolution in residual block is replaced by a ``$3\times3$ GConv followed by pointwise ($1\times1$) convolution''. We use various versions of WRN. WRN-22x2 represents the WideResNet with depth of 22 convolutional layers and 2 is the widening factor
by which number of filters in colnvolutional layers has increased. Similarly WRN-28x2, WRN-22x10, and WRN-40x2 is used for experiments. 
\item {\bf MobileNets}: We use MobileNetV2 for over-regularization experiments. 
\item {\bf ResNet}:  We use the ResNet-18 version where we replace the each $3\times3$ convolution in redsidual blocks with a  ``$3\times3$ GConv followed by pointwise ($1\times1$) convolution''
\end{itemize}

{\bf Effect of $g$, and $G$ on the performance of networks}
With increasing $g$, the number of channels in a group is decreased; hence each group captures fewer variations of a latent concept. This could lead to the hamper the representational power of the network. In standard convolution, each ofmap is connected to every ifmap in a fully-connected manner increases the filter-dependence. By contrast, in GConv, each ofmap is connected to only a group of ifmap; hence it acts as a regularizer that breaks the filter co-dependence and improves the generalization. The accuracy of all the baseline residual networks (students), except MobileNetV2, increases with higher number of channels per group (i.e, with increase in $G$ and decrease in $g$) (Table \ref{tab:ResNet18}, \ref{tab:WRN22x2}, \ref{tab:WRN22x10} \ref{tab:WRN28x2}, and \ref{tab:WRN40x2}). This shows that in these networks, the effect of an increase in representation dominates the effect of regularization. However,  for MobilenetV2, the accuracy trend is opposite, i.e., the accuracy of baseline residual networks increases with higher regularization 
(with an increase in $g$ and decrease in $G$) (Table \ref{tab:MV2}).  Hence, in MobileNetV2, the effect of regularization dominates the effect of representational power. Surprisingly, the accuracy of the baseline NR-G2  is higher than that of the NR-g2. We conjecture that the effect of regularization is higher in MobilenetV2 because it is a very deep but slim network.  

{\bf Role of residual connections}
\begin{table} [htbp]
\caption{Accuracy drop and distillation gain in WRN-22x2: students NR-g8, and NR-g16 in left table; NR-G2, and NR-G4 in right table achieve {\em noticeable positive gain in accuracy} through knowledge distillation as compared to their (respective) baseline models}
\label{tab:WRN22x2}
\begin{minipage}{.47\linewidth}
%\centering 
\resizebox{1.0\textwidth}{!}{
\begin{tabular} {c|c|c|c|c} 
\toprule
$I_t$-$S_s$ & NR-g2 & NR-g4 & NR-g8 & NR-g16 \\ \toprule
R-G1 & 66.52 & 68.08 & 69.42 & 70.56 \\
R-G2 & 66.68 & 68.57 & 69.56 & 69.38 \\
R-G4 & 66.46 & 69.02 & 69.6 & 70.96 \\
R-G8 & 66.65 & 68.13 & 70.64 & 69.05 \\
R-G16 & 66.54 & 67.86 & 70.42 & 70.32 \\ \midrule
Baseline & 71.13 & 69.92 & 69.29 & 68.74 \\
NR-baseline & 64.01 & 65.58 & 66.26 & 66.59 \\ \midrule
Acc. drop ($\downarrow$)  & 7.12  & 4.34 & 3.03 & 2.15 \\
Distil. gain ($\uparrow$) & -4.45 & -0.9 & {\bf 1.35} & {\bf 2.22} \\ \bottomrule
\end{tabular}}
\end{minipage}
\begin{minipage}{.53\linewidth}
%\centering
\resizebox{1.0\textwidth}{!}{
\begin{tabular} {c|c|c|c|c} 
\toprule
I.t-S.s & NR-G2 & NR-G4 & NR-G8 & NR-G16 \\ \toprule
R-g2 & 69.70 & 69.61 & 67.25 & 68.05 \\
R-g4 & 69.78 & 69.41 & 68.12 & 67.48 \\
R-g8 & 69.75 & 69.32 & 68.46 & 66.30 \\
R-g16 & 69.51 & 69.15 & 66.92 & 66.03 \\ \midrule
Baseline & 68.04 & 68.88 & 69.04 & 69.53 \\
NR-baseline & 65.98 & 65.82 & 65.19 & 65.08 \\ \midrule
Acc. drop ($\downarrow$) & 2.06 & 3.06 & 3.85 & 4.45 \\
Distil. gain ($\uparrow$) & {\bf 1.74} & {\bf 0.73} & -0.58 & -1.48 \\ \bottomrule
\end{tabular}}
\end{minipage}
\end{table} 
\begin{table} [htbp]
\caption{Accuracy drop and distillation gain in WRN-22x10: With  increase in width, grouped teachers get more benefits and hence all the students in right table  {\em achieve significant distillation gain.}}
\label{tab:WRN22x10}
\begin{minipage}{.47\linewidth}
%\centering 
\resizebox{1.0\textwidth}{!}{
\begin{tabular} {c|c|c|c|c} 
\toprule
$I_t$-$S_s$ & NR-g2 & NR-g4 & NR-g8 & NR-g16 \\ \toprule
R-G1 & 75.15 & 75.25 & 77.21 & 77.1 \\
R-G2 & 75 & 76.44 & 76.44 & 77.01 \\
R-G4 & 73.05 & 75.39 & 76.44 & 76.87 \\
R-G8 & 75.68 & 75.64 & 76.43 & 78.03 \\
R-G16 & 75.07 & 73.54 & 76.63 & 77.8 \\ \midrule
Baseline & 77.26 & 77.15 & 77.11 & 77.06 \\
NR-baseline & 69.61 & 71.05 & 71.83 & 73.32 \\ \midrule
Acc. drop ($\downarrow$) & 7.65 & 6.1 & 5.28 & 3.74 \\
Distil. gain ($\uparrow$) & -1.58 & -0.71 & {\bf 0.1} & {\bf 0.97} \\ \bottomrule
\end{tabular}}
\end{minipage}
\begin{minipage}{.53\linewidth}
%\centering
\resizebox{1.0\textwidth}{!}{
\begin{tabular} {c|c|c|c|c} 
\toprule
I.t-S.s & NR-G2 & NR-G4 & NR-G8 & NR-G16 \\ \toprule
R-g2 & 77.55 & 78.21 & 77.92 & 76.70 \\ 
R-g4 & 78.17 & 77.95 & 77.85 & 76.58 \\
R-g8 & 77.70 & 78.16 & 77.79 & 78.19 \\
R-g16 & 78.05 & 78.10 & 77.49 & 78.17 \\ \midrule
Baseline & 75.46 & 75.57 & 75.59 & 76.22 \\ 
NR-baseline & 74.63 & 74.16 & 73.7 & 71.1 \\ \midrule
Acc. drop ($\downarrow$)& 0.83 & 1.41 & 1.89 & 5.12 \\
Distil. gain ($\uparrow$) & {\bf 2.71} & {\bf 2.64} & {\bf 2.33} & {\bf 1.97} \\ \bottomrule
\end{tabular}}
\end{minipage}
\end{table} 
When all the residual connections are removed from baseline residual networks, the accuracy trend with increasing $g$/$G$ gets reversed except for WRN-40x2 and MobilenetV2 (Table \ref{tab:ResNet18}, \ref{tab:MV2}, \ref{tab:WRN22x2}, \ref{tab:WRN22x10} \ref{tab:WRN28x2}, and \ref{tab:WRN40x2}). This shows that residual connections in these networks serve the purpose of regularization, and once it is removed, higher regularized models get higher accuracy. However, for WRN-40x2, residual and non-residual baseline networks follow the accuracy trend in the order of representational power. This implies that the residual network in the baseline network serves the purpose of increasing representational power. As network (without residual) gets deeper, it incurs diminished feature reuse; however, with residual connections, it enable-feature reuse even in deeper network.  

{\bf Effect of width and depth}
As shown in (right) Table \ref{tab:WRN22x2}, only the students NR-G2 and NR-G4 have positive distillation gain. When the width of the networks increases, all the students in (right) Table \ref{tab:WRN22x10} have positive gain.  However, this effect does not reflected when the teachers have constant $G$ ((left) Table in \ref{tab:WRN22x2}, and Table \ref{tab:WRN22x10}). This happens because the teacher with constant $g$ gets more channels per group when width has increased; however, teachers with constant $G$ have the same number of channels even when width has increased. Thus, students distilled with teachers having a constant $g$ get benefit from increasing width.

When the depth of network increasing beyond a certain limit, the non-convexity of error surface gets worse, and there is almost no region with well-behaved lower loss values. Hence, merely with better initialization, one cannot recover the accuracy drop due to the removal of residual connections (Table \ref{tab:WRN40x2}). That is, {\em with higher depth, KD is not a good substitute for residual connections.}

\begin{table} [htbp]
\caption{Accuracy drop and distillation gain in WRN-28x2: students NR-g16 (left table) and NR-G2 (right table) {\em fully recover the drop in accuracy} through knowledge distillation}
\label{tab:WRN28x2}
\begin{minipage}{.47\linewidth}
%\centering 
\resizebox{1.0\textwidth}{!}{
\begin{tabular} {c|c|c|c|c} 
\toprule
I.t-S.s & NR-g2 & NR-g4 & NR-g8 & NR-g16 \\ \toprule
R-G1 & 66.02 & 65.63 & 68.66 & 70.02 \\
R-G2 & 62.43 & 65.38 & 67.06 & 69.95 \\
R-G4 & 66.43 & 65.4 & 67.28 & 68.94 \\
R-G8 & 64.62 & 65.5 & 68.71 & 68.87 \\
R-G16 & 65.01 & 65.8 & 64.98 & 69.6 \\ \midrule
Baseline & 70.84 & 70.3 & 69.93 & 69.57 \\
NR-baseline & 62.06 & 62.26 & 62.39 & 64.37 \\ \midrule
Acc. drop ($\downarrow$) & 8.78 & 8.04 & 7.54 & 5.2 \\
Distil. gain ($\uparrow$) & -4.41 & -4.5 & -1.22 & {\bf 0.45} \\ \bottomrule
\end{tabular}}
\end{minipage}
\begin{minipage}{.53\linewidth}
%\centering
\resizebox{1.0\textwidth}{!}{
\begin{tabular} {c|c|c|c|c} 
\toprule
I.t-S.s & NR-G2 & NR-G4 & NR-G8 & NR-G16 \\ \toprule
R-g2 & 68.98 & 67.32 & 66.56 & 65.24 \\
R-g4 & 68.72 & 68.37 & 64.37 & 65.49 \\
R-g8 & 69.14 & 68.05 & 67.42 & 65.45 \\
R-g16 & 68.94 & 68.52 & 67.66 & 65.39 \\ \midrule
Baseline & 68.62 & 68.82 & 69.14 & 70.28 \\
NR-baseline & 62.53 & 62.37 & 61.92 & 61.76 \\ \midrule
Acc. drop ($\downarrow$)& 6.09 & 6.45 & 7.22 & 8.52 \\
Distil. gain ($\uparrow$)& {\bf 0.52} & -0.3 & -1.48 & -4.79 \\ \bottomrule
\end{tabular}}
\end{minipage}
\end{table}

\begin{table} [htbp]
\caption{Accuracy drop and distillation gain in WRN-40x2: With  increase in depth accuracy drops are huge. Although students achieve a great boost in accuracy, {\em accuracy drops are not fully recovered}.}
\label{tab:WRN40x2}
\begin{minipage}{.47\linewidth}
%\centering 
\resizebox{1.0\textwidth}{!}{
\begin{tabular} {c|c|c|c|c} 
\toprule
$I_t$-$S_s$ & NR-g2 & NR-g4 & NR-g8 & NR-g16 \\ \toprule
R-G1 & 55.85 & 57.32 & 58.87 & 57.26 \\
R-G2 & 55.26 & 58.19 & 59.75 & 64.44 \\
R-G4 & 54.64 & 58.13 & 61.71 & 64.9 \\
R-G8 & 56.92 & 61.42 & 59.73 & 64.67 \\
R-G16 & 53.64 & 55.5 & 61.78 & 62.42 \\ \midrule
Baseline & 71.21 & 70.34 & 70.13 & 69.98 \\
NR-baseline & 43.92 & 42.86 & 42.58 & 40.49 \\ \midrule
Acc. drop ($\downarrow$) & 27.29 & 27.48 & 27.55 & 29.49 \\
Distil. gain ($\uparrow$) & -14.29 & -8.92 & -8.35 & -5.08 \\ \bottomrule
\end{tabular}}
\end{minipage}
\begin{minipage}{.53\linewidth}
%\centering
\resizebox{1.0\textwidth}{!}{
\begin{tabular} {c|c|c|c|c} 
\toprule
I.t-S.s & NR-G2 & NR-G4 & NR-G8 & NR-G16 \\ \toprule
R-g2 & 63.19 & 62.13 & 60.96 & 57.29 \\
R-g4 & 64.45 & 64.41 & 61.04 & 61.13 \\
R-g8 & 62.22 & 62.46 & 61.38 & 57.79 \\
R-g16 & 64.50 & 64.91 & 62.66 & 60.91 \\ \midrule
Baseline & 68.52 & 69.34 & 69.54 & 70.17 \\
NR-baseline & 38.47 & 45.88 & 47.55 & 49.49 \\ \midrule
Acc. drop ($\downarrow$) & 30.05 & 23.46 & 21.99 & 20.68 \\
Distil. gain ($\uparrow$) & -4.02 & -4.43 & -6.88 & -9.04 \\ \bottomrule
\end{tabular}}
\end{minipage}
\end{table}

\section{Visualization} \label{sec:Visualization}

\begin{figure}[htbp]
\centering
\begin{tabular}{cccc}
{\small Teacher n/w} &  {\small Baseline residual n/w}&  {\small Baseline non-residual n/w}&{\small Distilled n/w} \\ 
\includegraphics[scale=0.2]{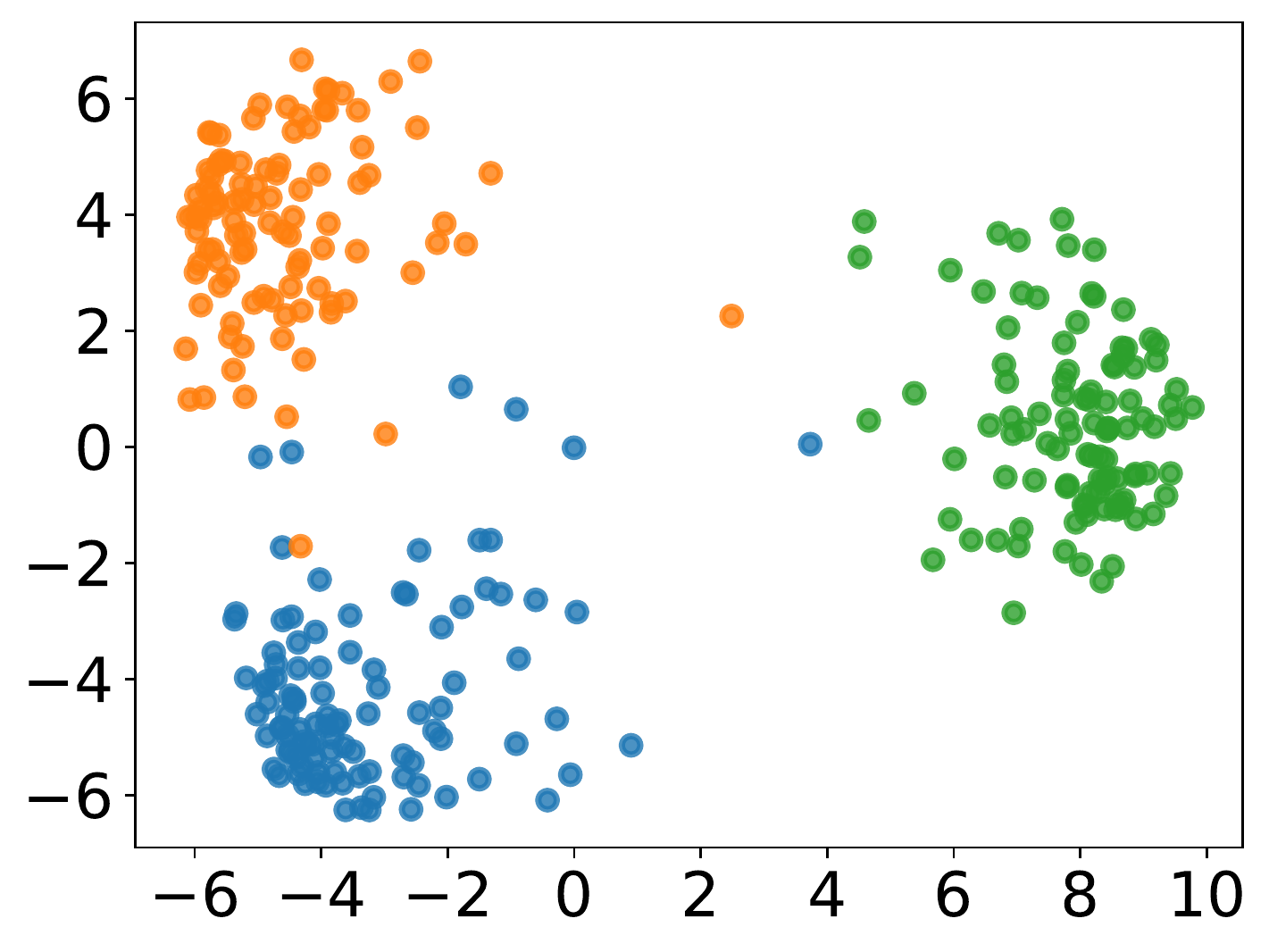}  & 
\includegraphics[scale=0.2]{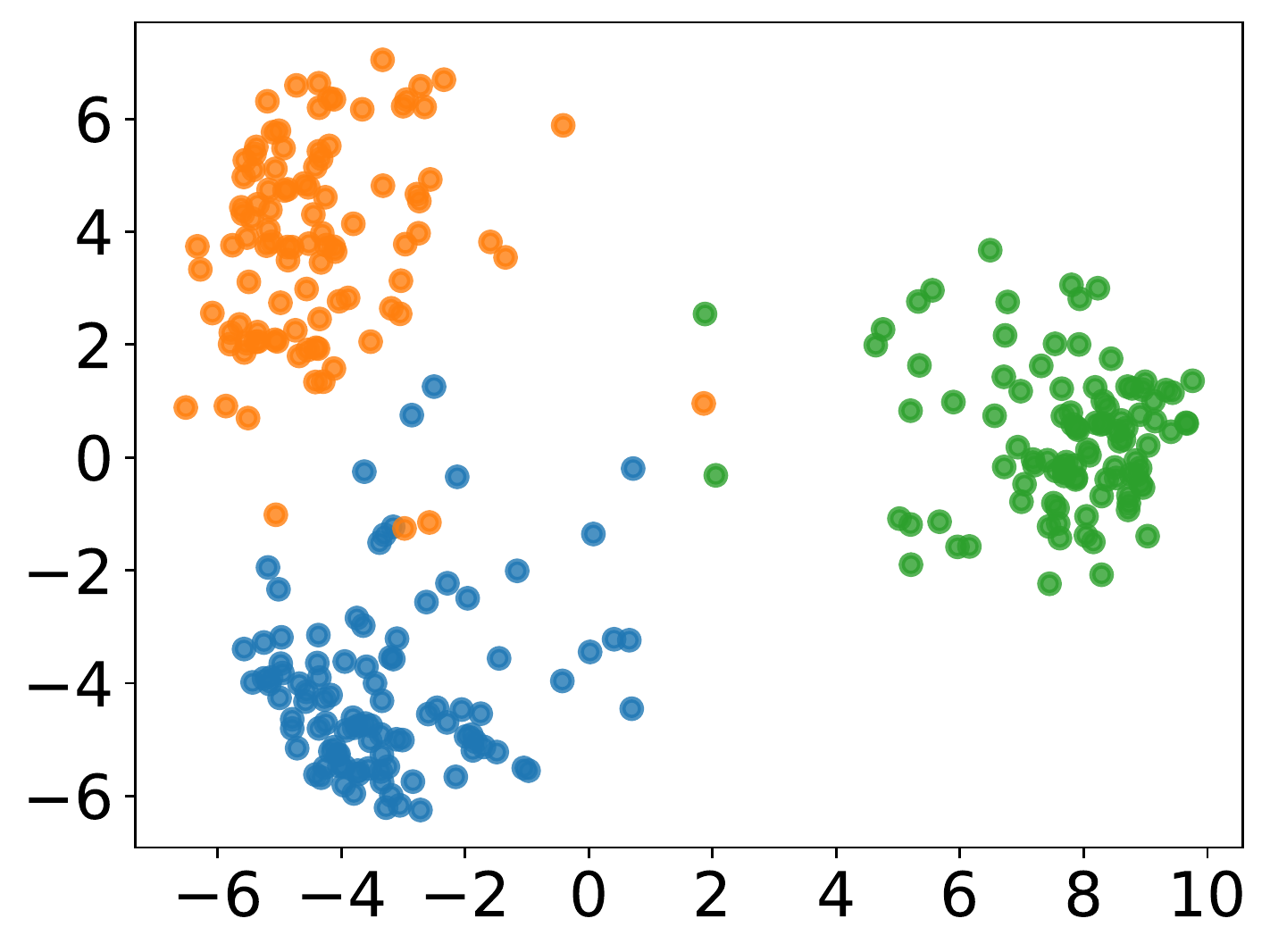}  & 
\includegraphics[scale=0.2]{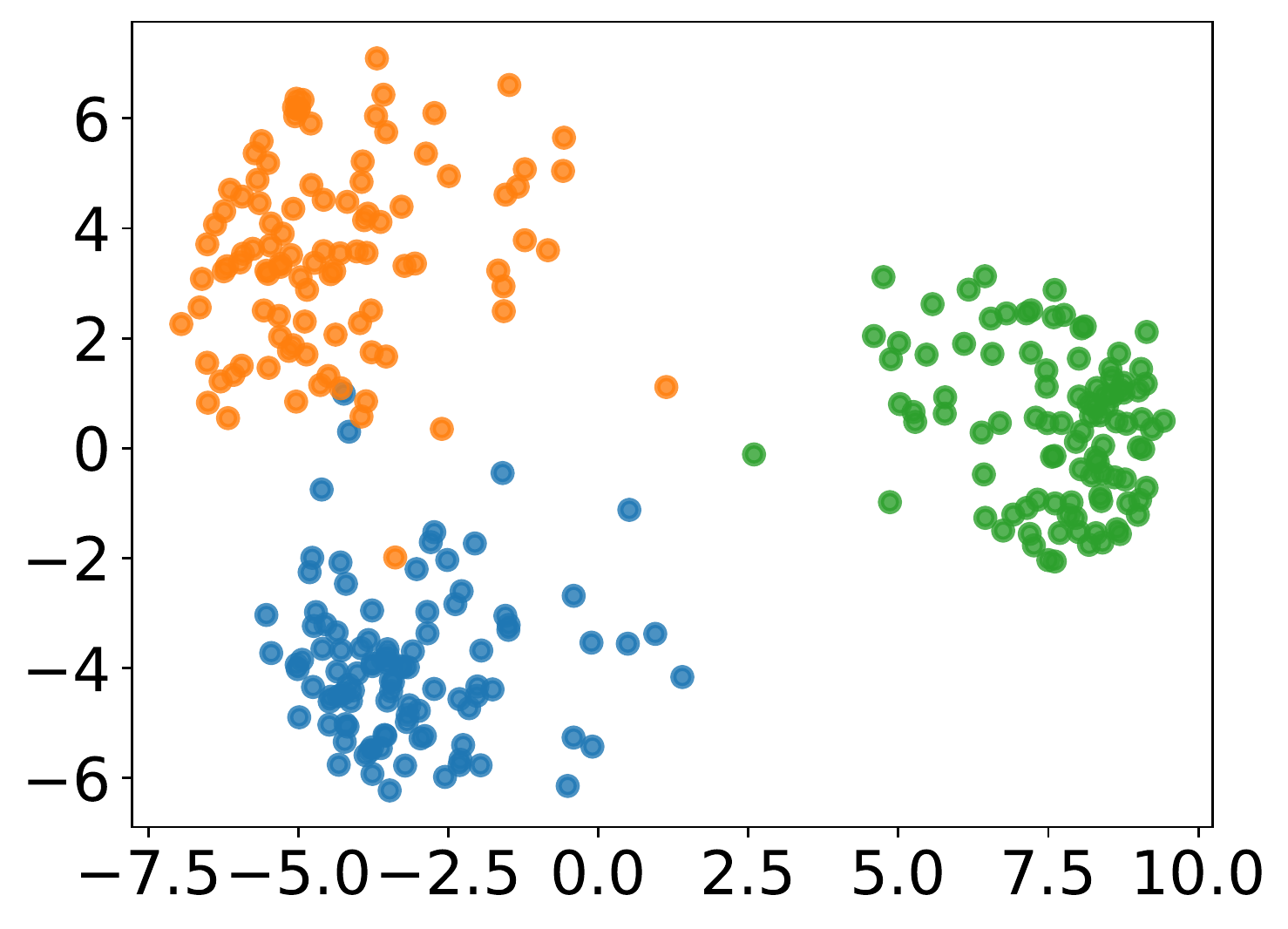}  &
\includegraphics[scale=0.2]{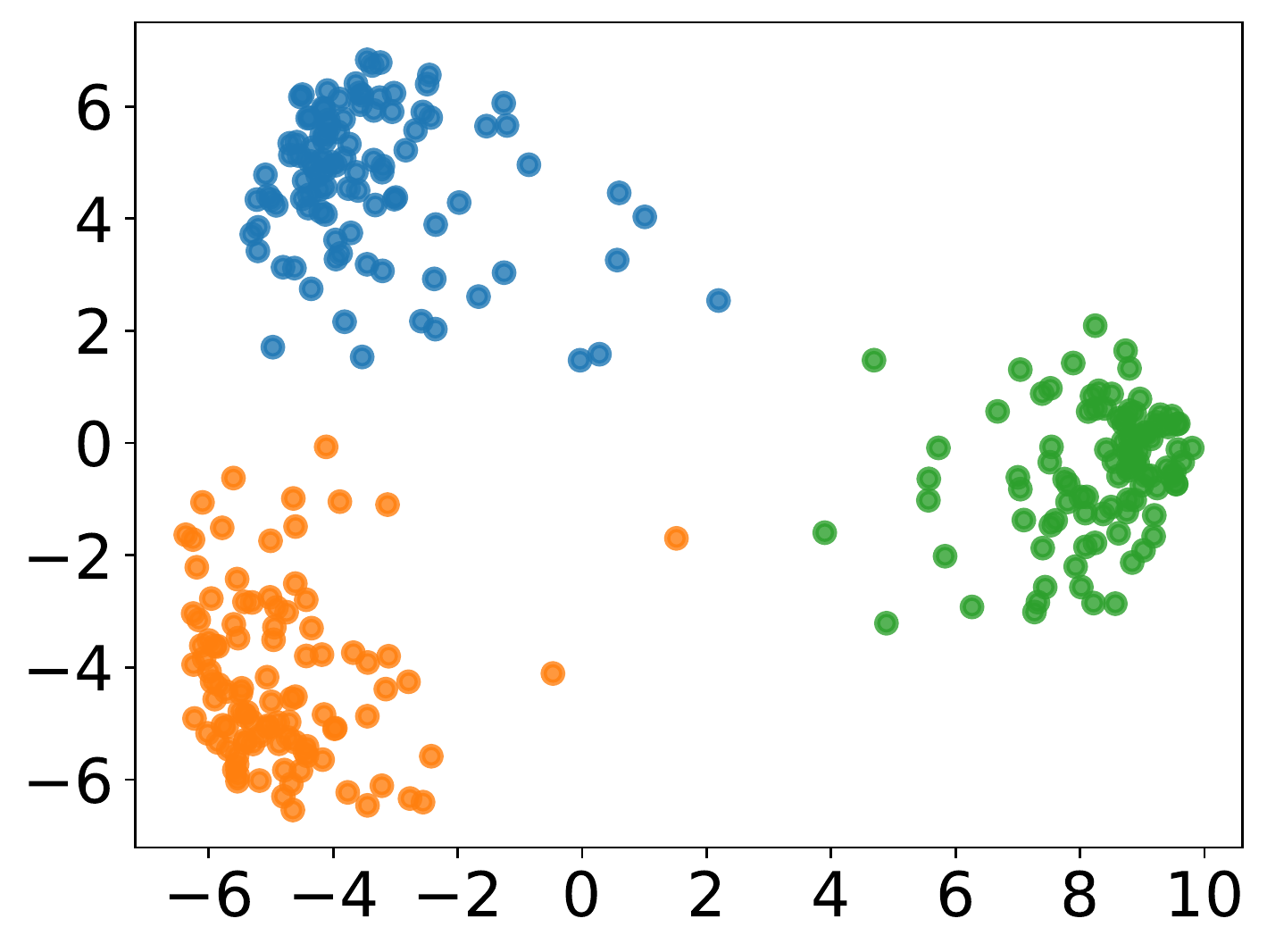} \\  
\includegraphics[scale=0.2]{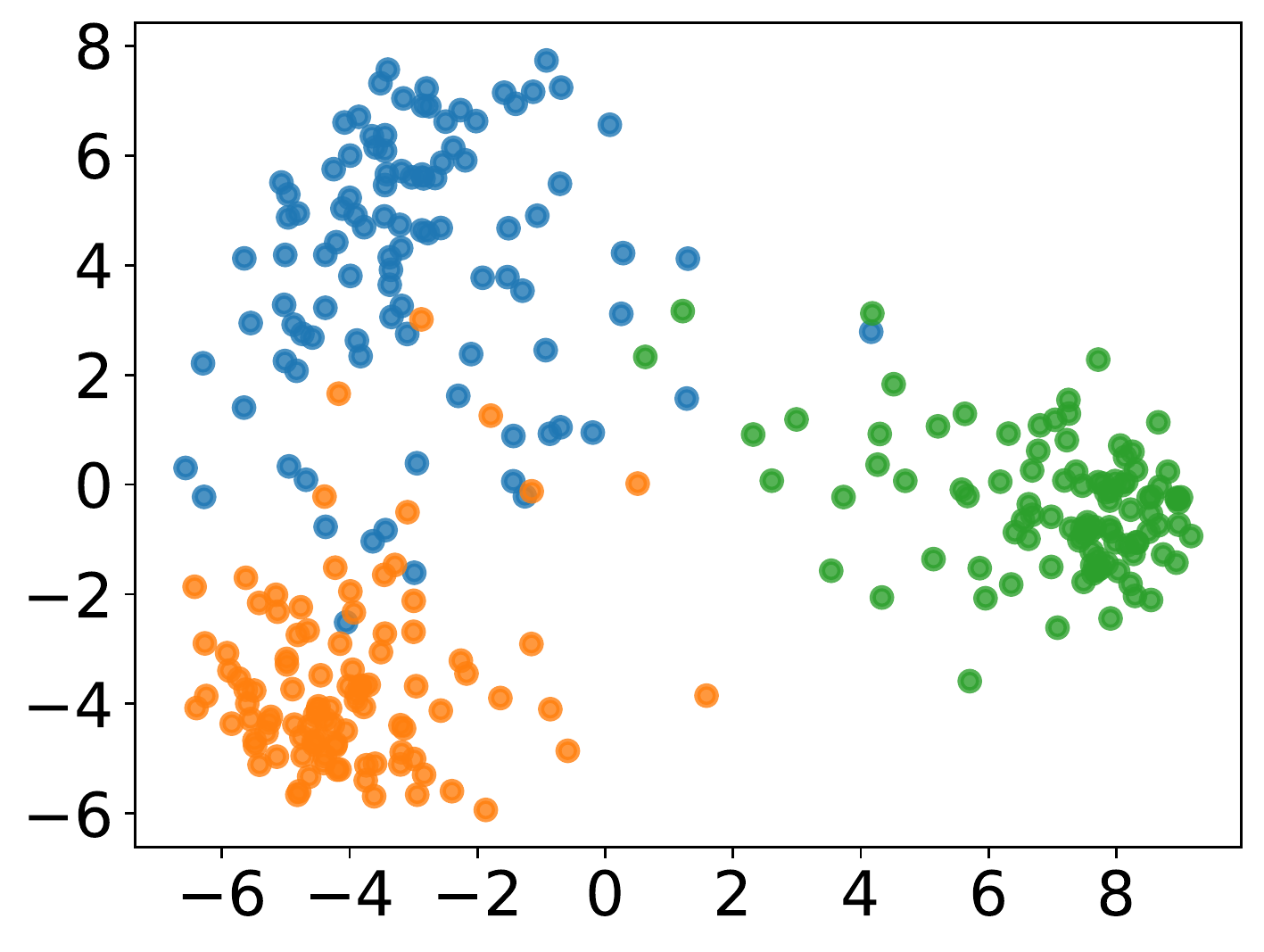}  &
\includegraphics[scale=0.2]{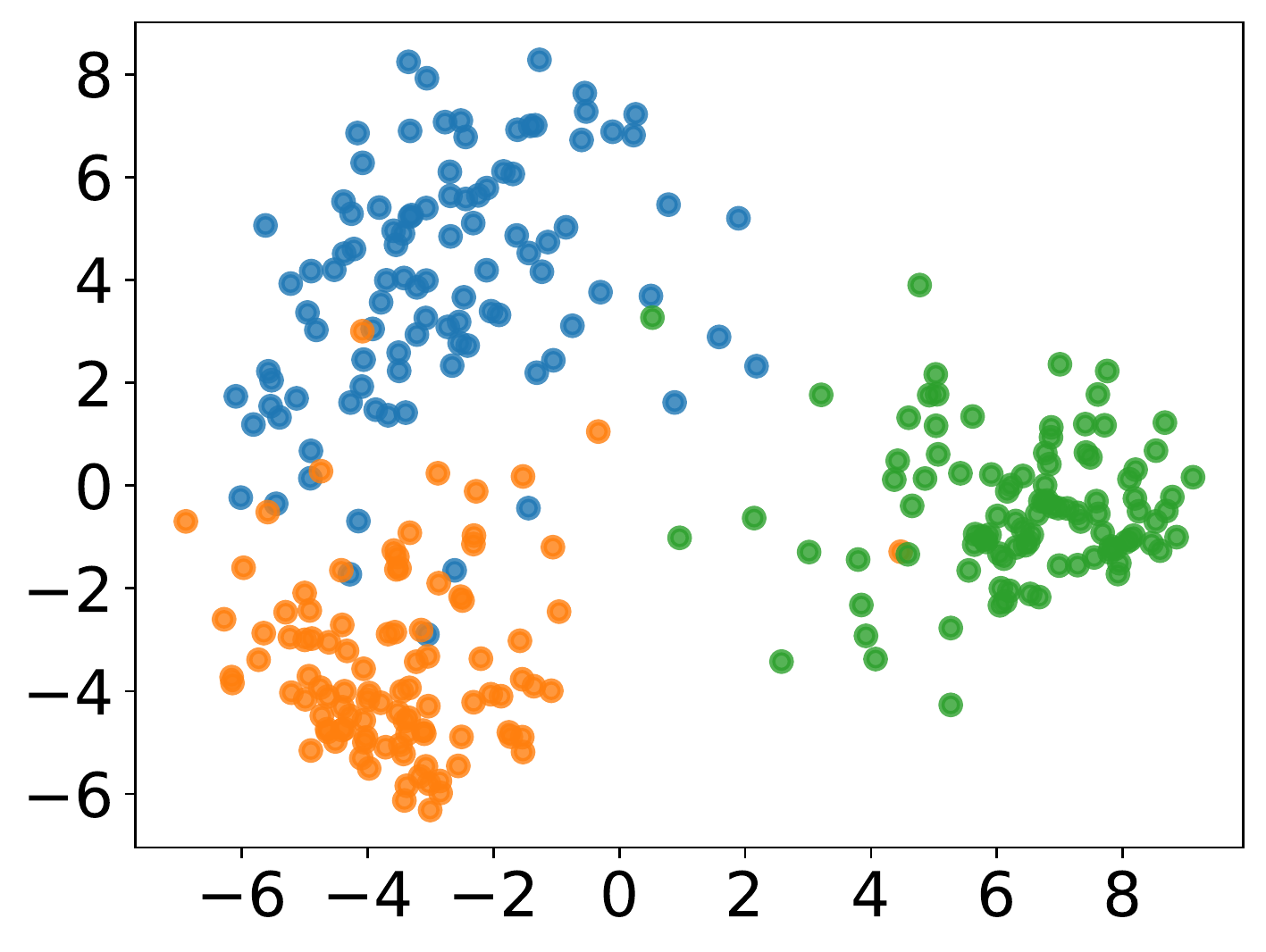}  &
\includegraphics[scale=0.2]{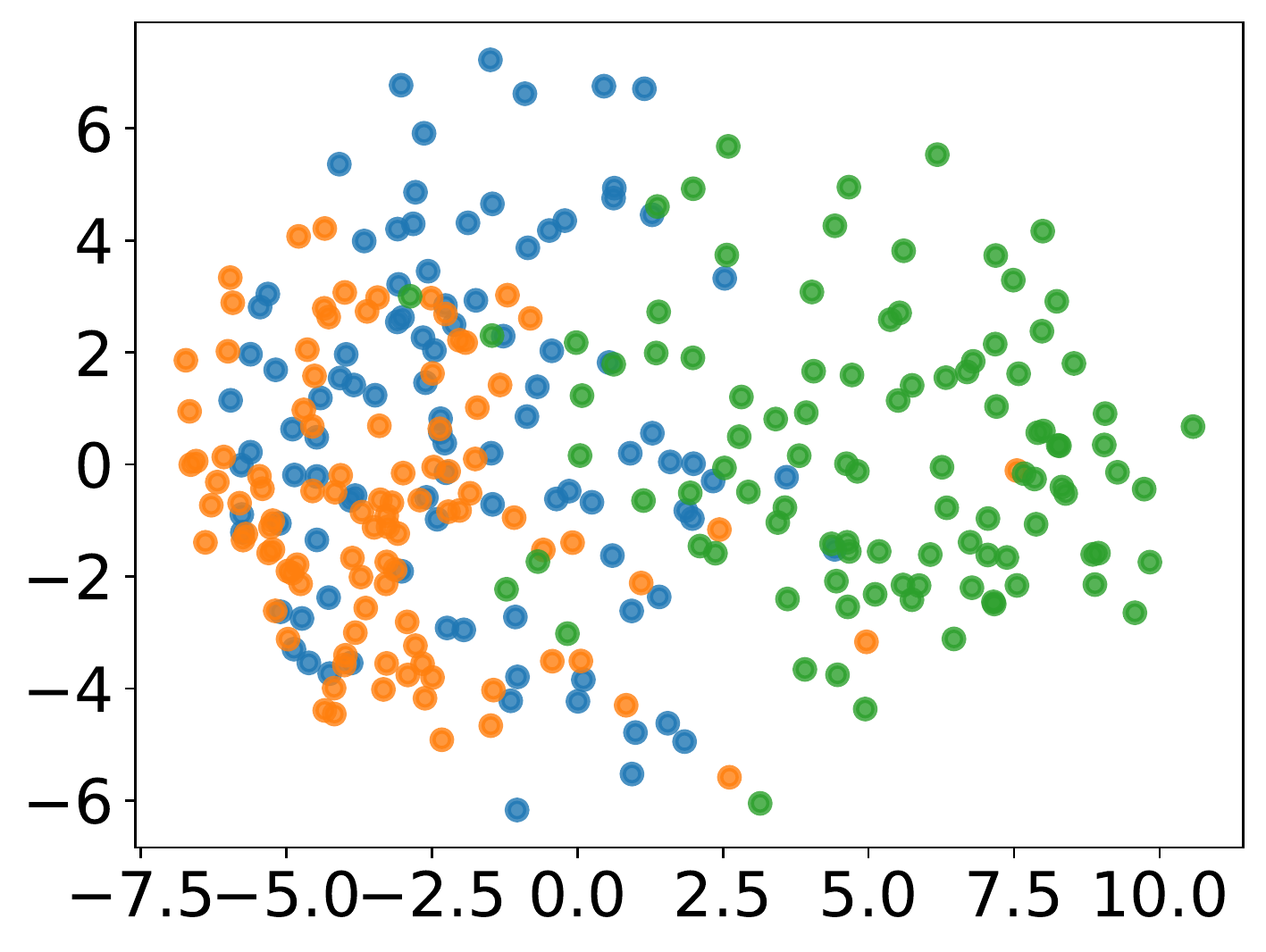}  & 
\includegraphics[scale=0.2]{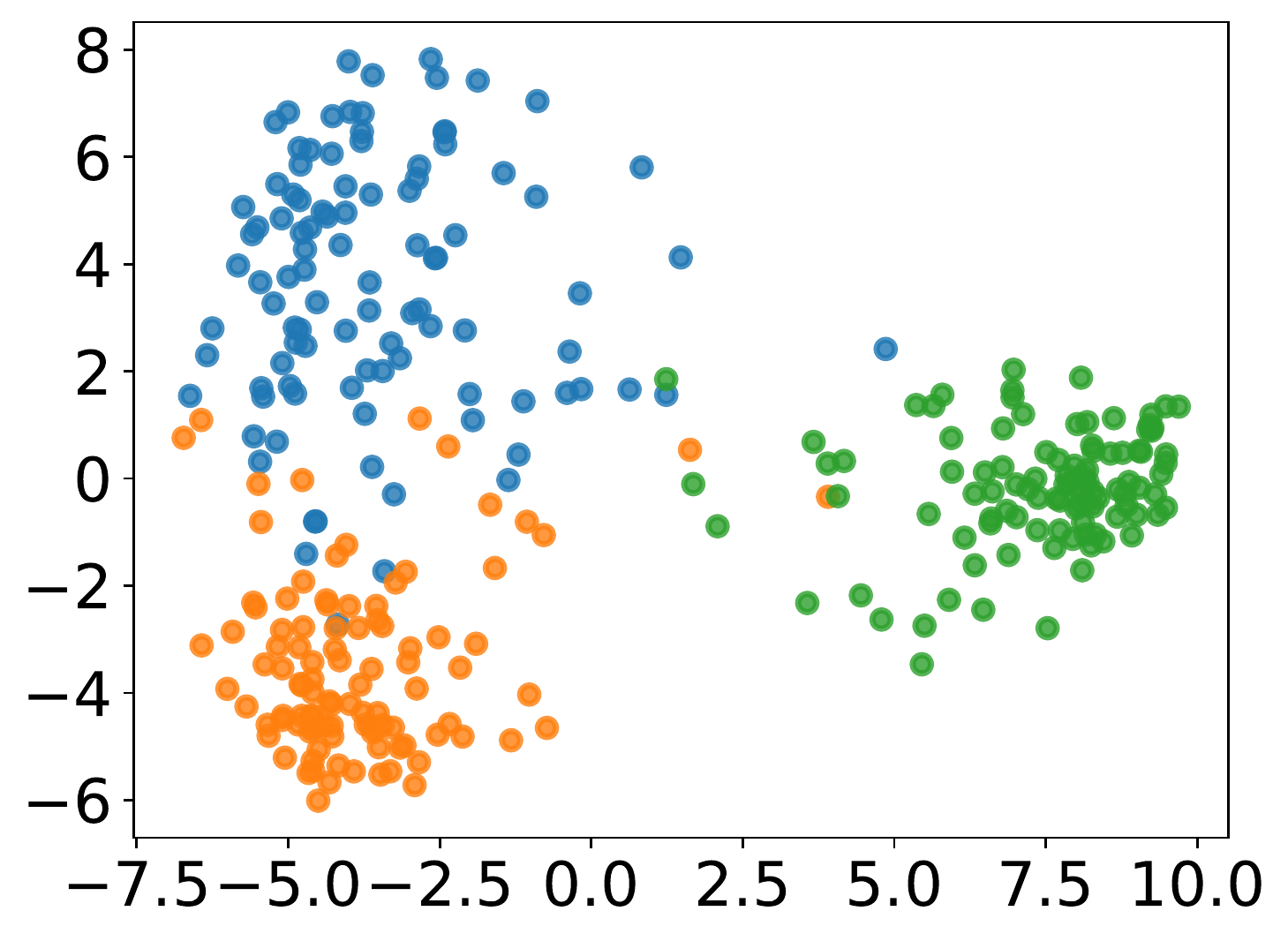}  \\ 
\includegraphics[scale=0.2]{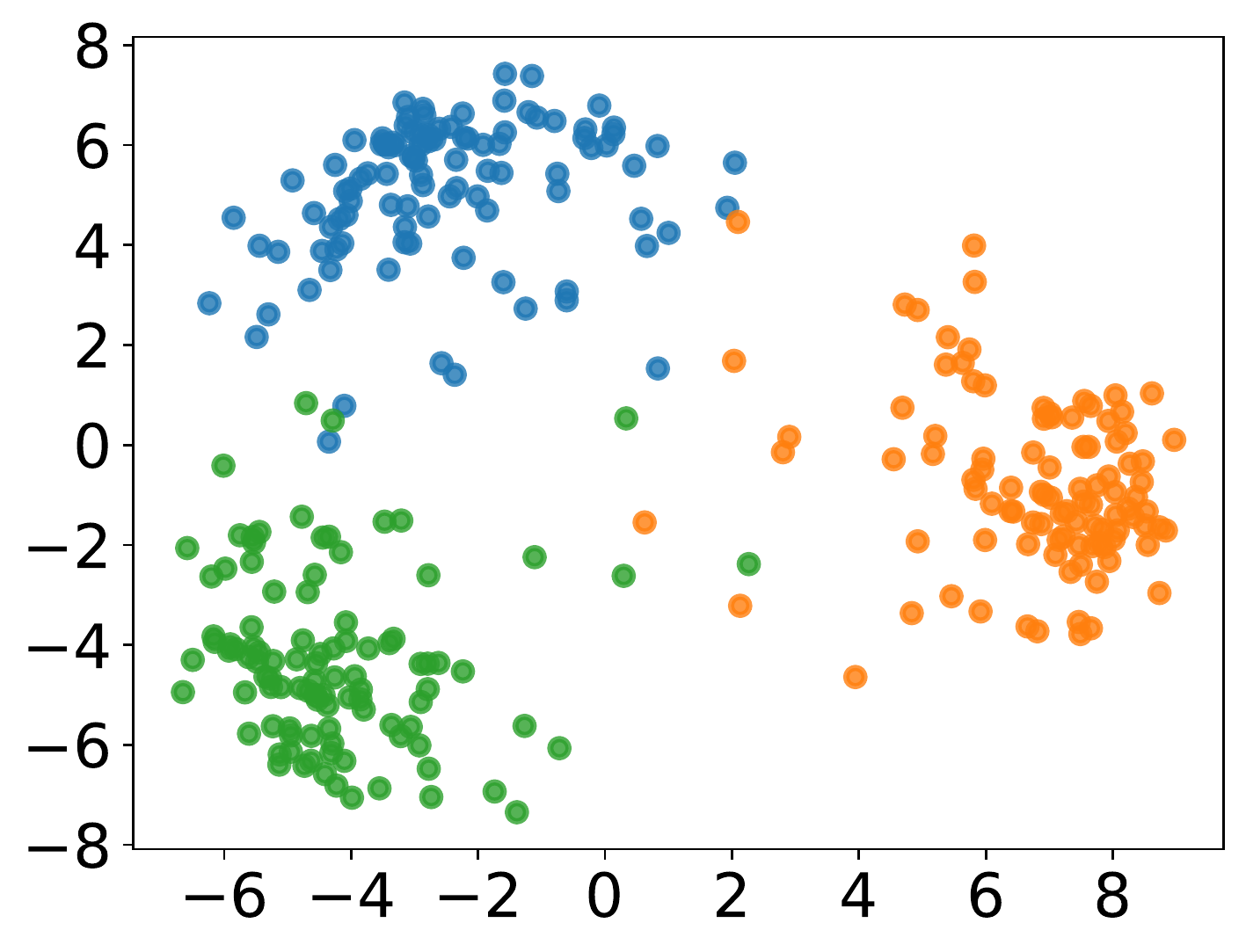}   &
\includegraphics[scale=0.2]{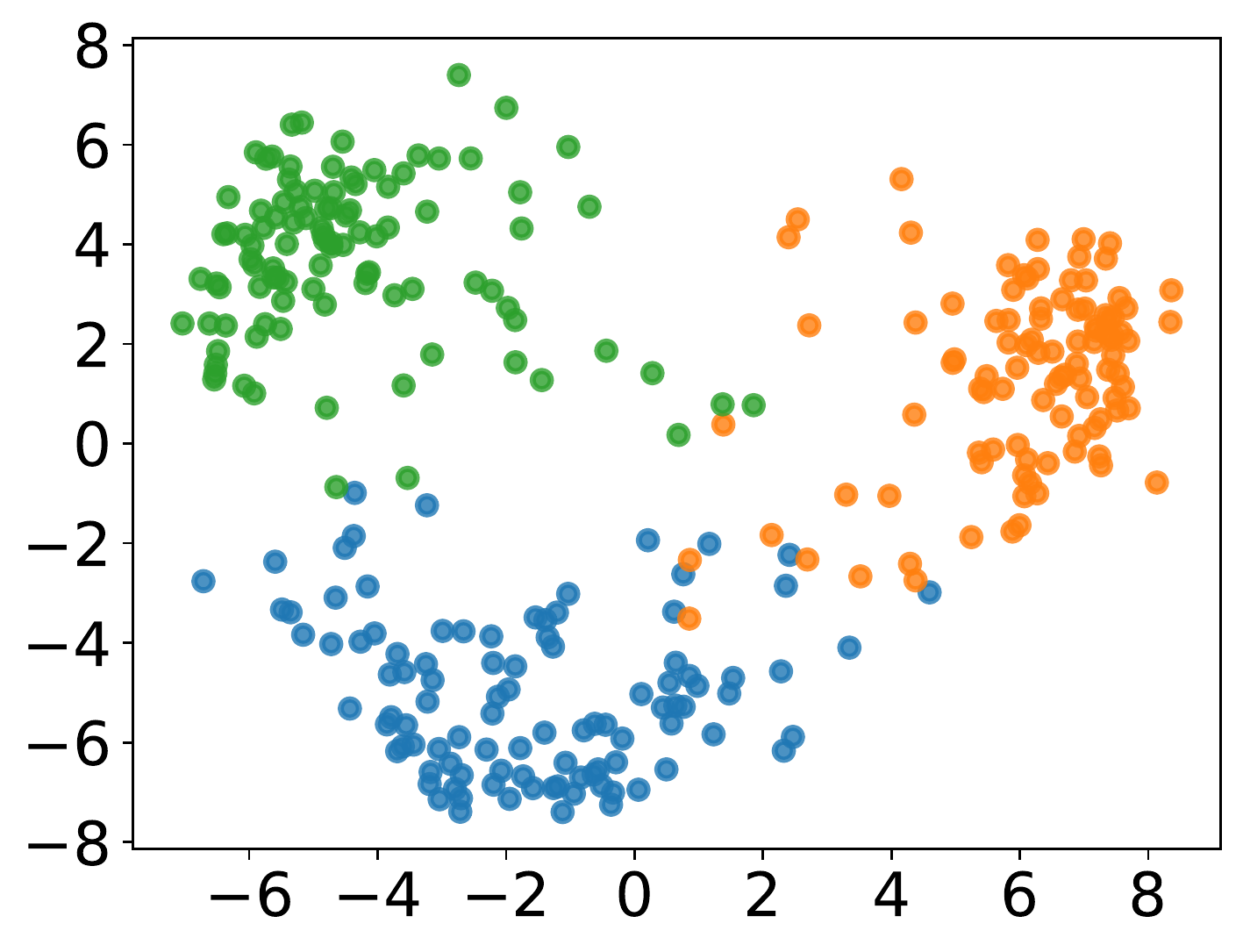}  & 
\includegraphics[scale=0.2]{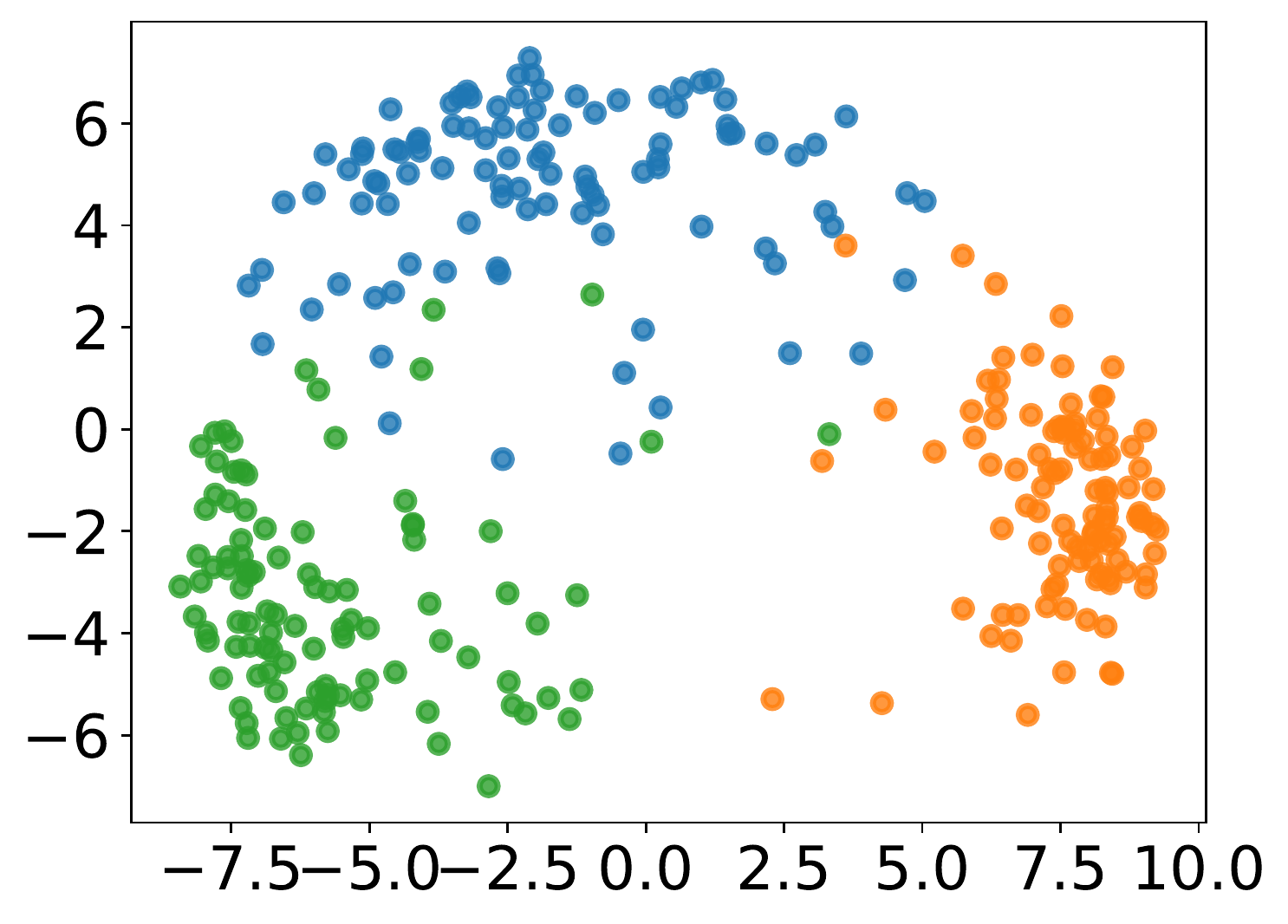}  & 
\includegraphics[scale=0.2]{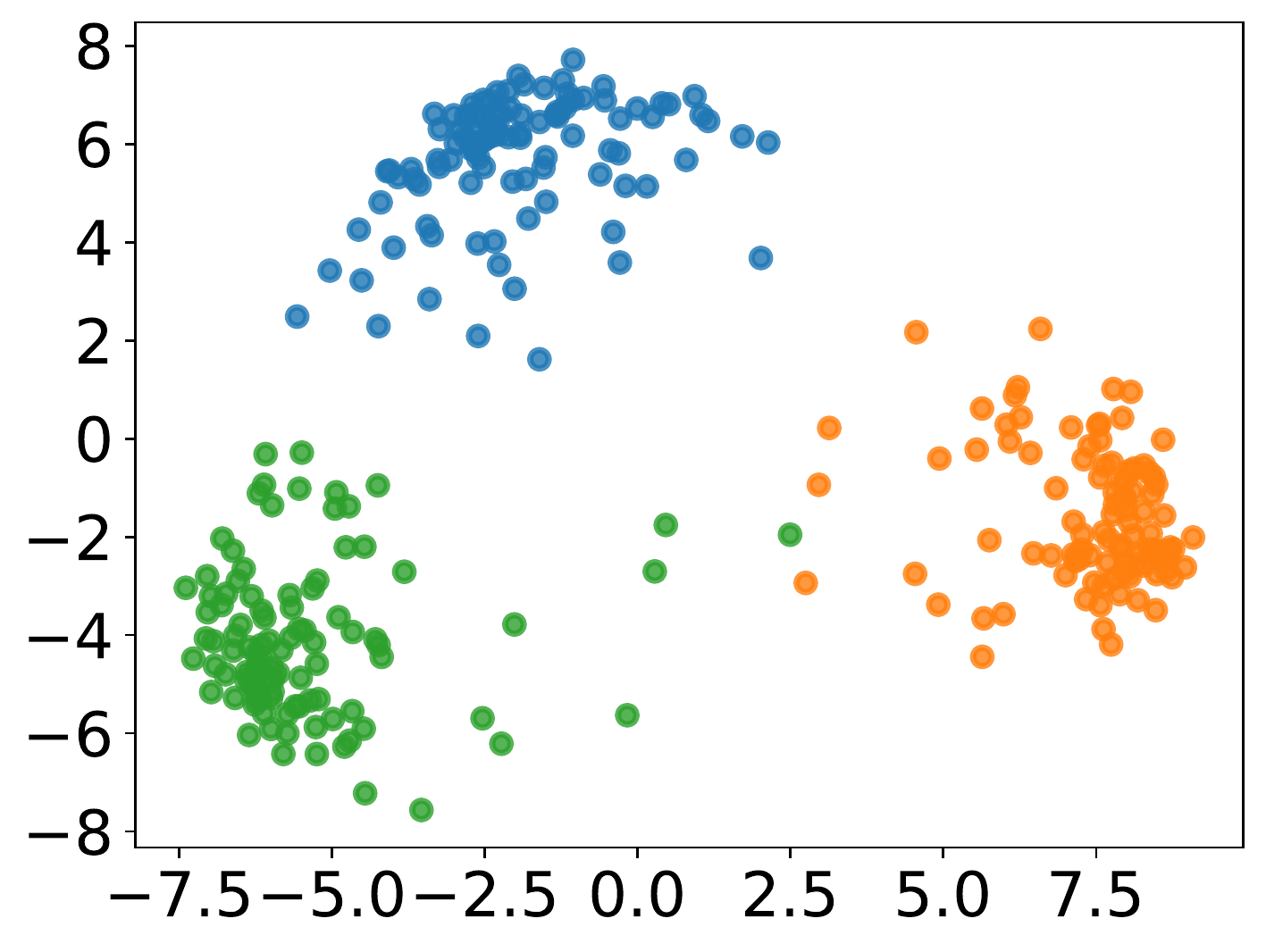} \\ 
\end{tabular}
\caption{Visualization for activations of last FC/conv layers: corresponding to three classes in CIFAR-100: upper row is WRN-22x10, middle row is WRN-40x2 and lower row is MobileNet-V2}
\label{fig:Visualization}    
\end{figure}

We use the visualizations for the activations of the last layer of WRN-22x10, WRN-40x2, and MobileNet-V2 (in order) to emphasize that KD can compensate for the accuracy drop due to the removal of residual connections from baseline network. We choose three semantically different classes on which the baseline non-residual network performed the worst (least top-1 accuracy). A cluster represents activations corresponding to these classes in Fig. \ref{fig:Visualization}. Well, separated clusters represent the network's ability to distinguish between classes while tight cluster represents the networks ability to bring the samples of the same class to its true distribution.

Top row (Fig. \ref{fig:Visualization}) visualizes the activations of WRN-22x10 on {\em House, Tractor} and {\em Whale} class (blue, orange, green respectively). Middle row visualizes the activations of WRN-40x2 on {\em Mushroom, Tiger} and {\em Train} class (blue, orange, green respectively). Last row represents the activations of MobileNet-V2 for {\em Bed}, {\em Mouse} and {\em Train} class (blue, orange, green respectively). The teacher network (column 1 in Fig. \ref{fig:Visualization}) effectively distinguish between the classes as clusters are well contained and separate from each other except few points. The baseline residual model, also, can distinguish these classes very-well as clusters are well-separated from each other. However, in some cases, the student model without residual connection (col 3 in Fig. \ref{fig:Visualization}) is not able to distinguish well the semantically different classes as the clusters are broad and overlapping. These drawbacks are more pronounced in the deeper networks such as WRN40x2 (row 2 col 3 in Fig. \ref{fig:Visualization}) and MobileNet-V2 (row 3, col 3 in Fig. \ref{fig:Visualization}). 
This highlights the {\em  importance of residual connection in deeper networks}. The clusters (row 2, col 4) are much tighter as compared to baseline non-residual network. In MobileNet-V2 (row 3 and col 4), the clusters are tight and well separated.

Distilled WRN-22x10 (row 1, col 4 in Fig. \ref{fig:Visualization}) outperform even the teacher network. The pictorial illustration shows that clusters are much tighter than that of the teacher and baseline networks.  This confirms that KD can compensate for the performance loss due to the removal of residual up to a great extent.  Thus, KD encourages the student to use the teacher's hint and helps in maximizing the inter-cluster distance. This further substantiates our claims that the KD can compensate for the residual connection.
\section{Conclusion}

In this work, we demonstrated the efficacy of knowledge distillation as a substitute for residual connections in the shallower network. We found that non-residual student networks not only recover the accuracy drop, which is incurred by the removal of all the residual connections; in some cases, it surpasses the accuracy of both the baseline residual network and identically structured teacher network. We show that KD enables a better initialization, which leads to convergence into the non-chaotic region in the error surface.

%\section{Broader Impact}
%
%Since this is works improves the understanding of knowledge distillation and presents a new perspective which does not lead to any SOTA performance, the broader impact is not applicable.  

%\section*{References}

{\small
\bibliographystyle{ieee}
\bibliography{References}
}

\newpage
\appendix

\section{Parameters and FLOPs in Residual Networks}

\subsection{Calculation of Parameters and FLOPs for Different Types of Convolution}
Let  input tensor $X$ $\in$ $\mathbb{R}^{m\times f\times f}$ is transformed into output tensor $Y$ $\in$ $\mathbb{R}^{n\times f\times f}$  by employing (standard)convolution using filters  $F$ $\in$ $\mathbb{R}^{n\times m\times k\times k}$ with appropriate padding and strides. The number of computations (measures as FLOPs) and the number of parameters involved in this transformation via standard convolution are shown in Table \ref{tab:ConvComp}.  The term $n\times m$ in the computation results in a very high number of FLOPs in deeper layers where $n$ and $m$ are higher (e.g., 512, 1024, 2048). To mitigate the high computational overhead in standard convolution (SConv), depthwise convolution (DWConv) has been proposed, which reduces the computation as well as the number of parameters by a factor of $n$ (Table \ref{tab:ConvComp}). However, it incurs fragmented memory-accesses due to one to one convolution between filter channels and input feature maps. This significantly lowered the advantage of reduced computation in DWConv, and the actual wall clock time for latency got worsened.

\begin{table} [htbp] 
\caption{Comparison of the number of parameters and computations in different types of convolution} 
\label{tab:ConvComp}\centering 
\resizebox{1.0\textwidth}{!}{
\begin{tabular}{cccccccc} \toprule
\multicolumn{2}{c}{ Standard conv } & \multicolumn{2}{c}{ GConv with constant $g$ } & \multicolumn{2}{c}{ GConv with constant $G$ } & \multicolumn{2}{c}{ Depthwise conv } \\ 
\#FLOPs & \#Param & \#FLOPs & \#Param & \#FLOPs & \#Param & \#FLOPs & \#Param \\ \toprule
$n\times m\times  k^2\times f^2$ & $n\times m\times k^2$ & $\frac{n\times m\times  k^2\times f^2}{g}$ & $\frac{n\times m\times k^2}{g}$ & $G\times(n\times f^2\times k^2)$ & $G\times(n\times k^2)$ & $m\times  k^2\times f^2$ & $m\times k^2$  \\ \bottomrule
\end{tabular} }
\end{table}

To enable a sweet point between standard convolution and depthwise convolution, group convolution (GConv) has been proposed. We implement two versions of GConv, first when the number of groups ($g$) remains constant in all the layers of a network, and second when the number of channels ($G$) in a group remains constant throughout the network. In the former, the number of channels in groups (indicated as the width of the individual boxes in Figure \ref{fig:GroupIllustration}(a)) keep growing in deeper layers, whereas the number of groups (indicated as the number of boxes in Figure \ref{fig:GroupIllustration}(b)) keep increasing in the deeper layers (since $g\times G=m$). In GConv input feature maps and filter's channels are divided into $t$ non-overlapping groups such that $X$ = \{$X_1$, $X_2$, ..., $X_t$\} and $F$ = \{$F_1$, $F_2$, ..., $F_t$\} where each groups have $\frac{m}{t}$ feature-maps. Note that $t$ for GConv with constant $g$ is $g$ and that for GConv with constant $G$ is $\frac{m}{t}$. This division of groups reduces the computations and the number of parameters, as shown in Table \ref{tab:ConvComp}. In GConv with constant $g$, both the number of parameters and FLOPs reduced by a factor of $g$. Compared to DWConv, the number of parameters and FLOPs in GConv with constant $G$ increases by a factor of $G$ and reduces the fragmented memory accesses in DWConv.

\begin{figure}[htbp]
\centering
\includegraphics[scale=0.5]{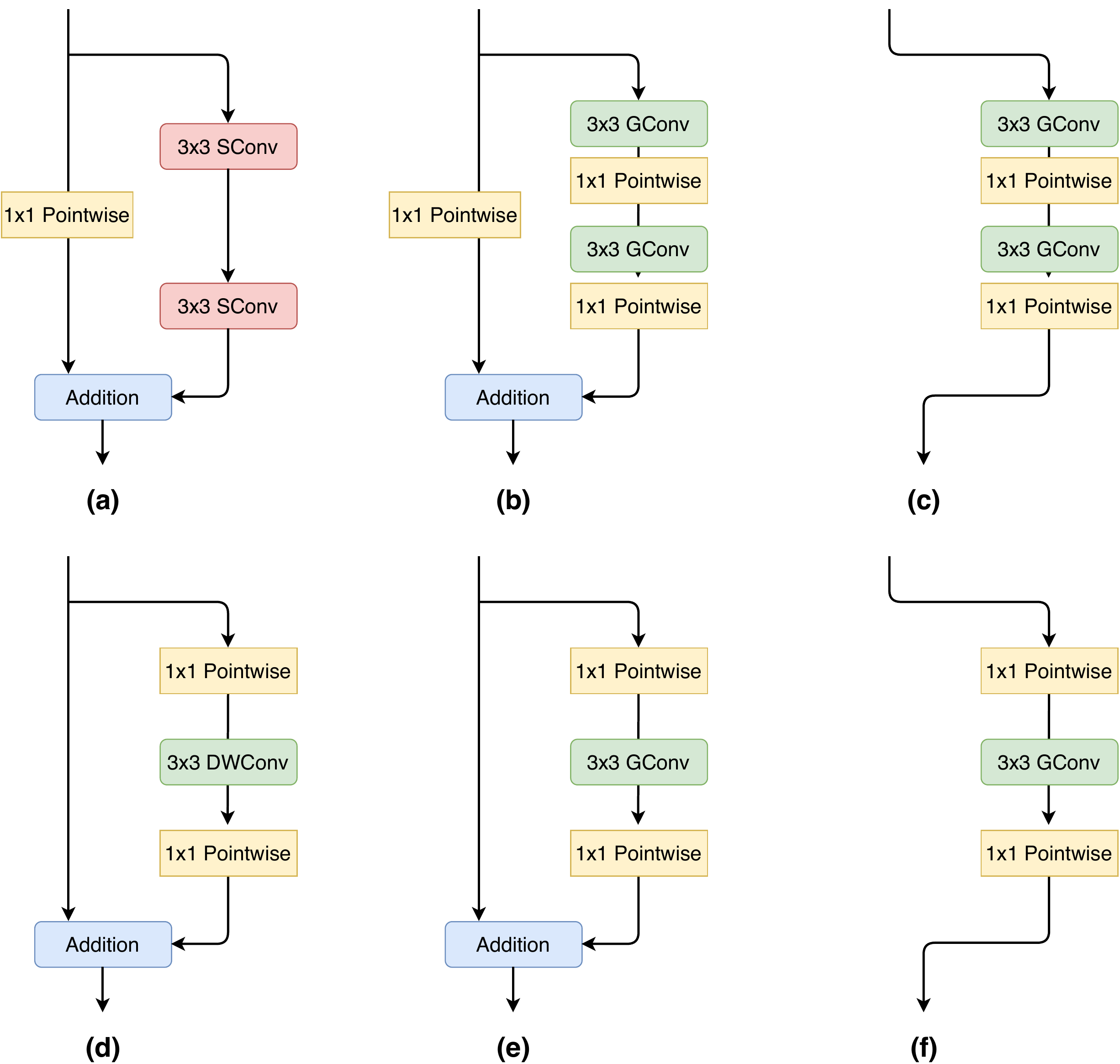} 
\caption{Residual block diagrams: Upper row corresponds to the residual blocks in WideResNet variants (and ResNet-18) and lower row is for MobileNet-V2. (a) Original residual blocks in WideResNet variants (and ResNet-18) have two $3\times3$ standard convolution (SConv). (b) Modified residual blocks for WideResNet variants and ResNet-18 where we replace each $3\times3$ with $3\times3$ GConv. (c) Non-residual version of modified residual blocks. (d) Original building blocks in MobileNet-V2. (e) Modified building blocks where we replace $3\times3$ DWConv with $3\times3$ GConv. (f) Non-residual version of modified building block.}
\label{fig:ResidualBlocks}
\end{figure}

\subsection{Comparison of Parameters and FLOPs in Residual Networks}
We leverage the idea of GConv in our experiments, where we create models with similar architecture but varying complexity by changing $g$/$G$. We replace each $3\times3$ standard convolution in the residual blocks of both WideResNet variants \cite{zagoruyko2016wide} and ResNet-18 \cite{he2016deep} with ``$3\times3$ GConv (with constant $g$/$G$) followed by $1\times1$ convolution'' as shown in upper row of Figure \ref{fig:ResidualBlocks}. The $1\times1$ convolution is deployed to blend the information from different groups of output feature maps (produced by GConv) and prevents the accuracy drop since, in GConv, information is taken only from a part of the input. Similarly, in the  original building blocks of MobileNet-V2 \cite{sandler2018mobilenetv2} we replace  $3\times3$ depthwise convolution layer with $3\times3$ GConv (with constant $g$/$G$) (lower row in Figure \ref{fig:ResidualBlocks}). We do not need to put an extra $1\times1$ convolution layer since it is already present in the original building blocks of MobileNet-V2. 
Note that MobileNet-V2 has inverted residual connections where the first layer in building block is $1\times1$ convolution, unlike WideResNet and ResNet-18, where the first convolution layer in the residual block is $3\times3$. Also, unlike WideResNet, residual connections in MobileNet-V2 do not have $1\times1$ convolution.

\begin{table} [htbp]
\caption{Number of FLOPs in Millions. Increasing $g$ results in decreases in FLOPs as it approaches towards depthwise convolution. Increasing $G$ increases FLOPs as it approaches towards standard convolution. Depthwise variants ($G$=1) has minimum number of FLOPs. Note that residual connections in all the networks, except MobileNet-V2, have $1\times1$ convolution, hence with its removal computation decreases. R and NR are residual and non-residual version of networks, respectively.}
\label{tab:FLOPsSummary}
\begin{tabular}{ccccccccccc} \toprule
Network & & g2 & g4 & g8 & g16 & G1 & G2 & G4 & G8 & G16 \\ \toprule
\multirow{2}{*}{ WRN-22x2 } & R & 93.65 & 56.49 & 37.91 & 28.62 & 22.17 & 25.01 & 30.68 & 42.04 & 64.75 \\
& NR & 92.07 & 54.92 & 36.34 & 27.05 & 20.6 & 23.43 & 29.11 & 40.47 & 63.17 \\ \midrule
\multirow{2}{*}{ WRN-22x10 } & R & 2281 & 1366 & 908.8 & 680.1 & 465 & 478.6 & 505.8 & 560.2 & 669 \\
& NR & 2252 & 1337 & 880 & 651.3 & 436.2 & 449.8 & 477 & 531.4 & 640.2 \\ \midrule
\multirow{2}{*}{ WRN-28x2 } & R & 128.3 & 76.94 & 51.28 & 38.45 & 29.49 & 33.36 & 41.1 & 56.59 & 87.55 \\
& NR & 126.7 & 75.36 & 49.71 & 36.88 & 27.92 & 31.79 & 39.53 & 55.01 & 85.98 \\ \midrule
\multirow{2}{*}{ WRN-40x2 } & R & 197.5 & 117.8 & 78.02 & 58.11 & 44.14 & 50.07 & 61.94 & 85.68 & 133.2 \\
& NR & 195.9 & 116.3 & 76.44 & 56.54 & 42.57 & 48.5 & 60.37 & 84.11 & 131.6 \\ \midrule
\multirow{2}{*}{ ResNet-18 } & R & 328.5 & 198.7 & 133.8 & 101.4 & 73.1 & 77.26 & 85.59 & 102.3 & 135.6 \\
& NR & 322.2 & 192.4 & 127.5 & 95.08 & 66.8 & 70.97 & 79.3 & 95.96 & 129.3 \\ \midrule
\multirow{2}{*}{ MobileNet-V2 } & R & 634.4 & 348.3 & 205.2 & 133.7 & 64.95 & 67.72 & 73.25 & 84.32 & 106.5 \\
& NR & 634.4 & 348.3 & 205.2 & 133.7 & 64.95 & 67.72 & 73.25 & 84.32 & 106.5 \\ \bottomrule
\end{tabular} 

\end{table}
\begin{table} [htbp]
\caption{Number of parameters in Millions. Increasing $g$ results in a reduction in the number of parameters  as it approaches towards depthwise convolution. Increasing $G$ increases the number of parameters as it approaches towards standard convolution. Depthwise variants ($G$=1) has minimum number of parameters. Notice that residual connections in all the networks, except MobileNet-V2, have $1\times1$ convolution, hence with its removal number of parameters decreases. R and NR are residual and non-residual version of networks, respectively.}
\label{tab:ParamsSummary}
\begin{tabular}{ccccccccccc} \toprule
Networks & & g2 & g4 & g8 & g16 & G1 & G2 & G4 & G8 & G16 \\ \toprule
\multirow{2}{*}{ WRN-22x2 } & R & 0.656 & 0.402 & 0.275 & 0.211 & 0.159 & 0.17 & 0.192 & 0.236 & 0.325 \\
& NR & 0.645 & 0.391 & 0.264 & 0.2 & 0.148 & 0.159 & 0.181 & 0.225 & 0.314 \\ \midrule
\multirow{2}{*}{ WRN-22x10 } & R & 16 & 9.63 & 6.46 & 4.88 & 3.35 & 3.41 & 3.51 & 3.73 & 4.17 \\
& NR & 15.7 & 9.37 & 6.21 & 4.62 & 3.09 & 3.15 & 3.26 & 3.48 & 3.91 \\ \midrule
\multirow{2}{*}{ WRN-28x2 } & R & 0.894 & 0.543 & 0.368 & 0.28 & 0.207 & 0.223 & 0.253 & 0.313 & 0.434 \\
& NR & 0.883 & 0.532 & 0.357 & 0.269 & 0.197 & 0.212 & 0.242 & 0.303 & 0.424 \\ \midrule
\multirow{2}{*}{ WRN-40x2 } & R & 1.37 & 0.826 & 0.554 & 0.418 & 0.305 & 0.328 & 0.375 & 0.467 & 0.653 \\
& NR & 1.36 & 0.816 & 0.543 & 0.407 & 0.294 & 0.318 & 0.364 & 0.457 & 0.642 \\ \midrule
\multirow{2}{*}{ ResNet-18 } & R & 6.57 & 4.01 & 2.74 & 2.1 & 1.49 & 1.52 & 1.58 & 1.71 & 1.95 \\
& NR & 6.39 & 3.84 & 2.56 & 1.93 & 1.32 & 1.35 & 1.41 & 1.53 & 1.78 \\ \midrule
\multirow{2}{*}{ MobileNet-V2 } & R & 22.6 & 12.5 & 7.38 & 4.84 & 2.37 & 2.43 & 2.56 & 2.82 & 3.33 \\
& NR & 22.6 & 12.5 & 7.38 & 4.84 & 2.37 & 2.43 & 2.56 & 2.82 & 3.33 \\ \bottomrule
\end{tabular} 

\end{table}

We have calculated the number of parameters, and the number of FLOPs for different versions of WideResNet, ResNet-18, and MobileNet-V2 in the Table \ref{tab:ParamsSummary} and Table  \ref{tab:FLOPsSummary}, respectively.  Notice that the number of computations (FLOPs) depends on the feature map size; hence it varies across datasets of different spatial size inputs. In other words, the same model incurs higher computations when it processes with a dataset of higher spatial resolution inputs. For example, a model will have significantly higher computations when it is used with the ImageNet-1K dataset (input resolution is $224\times 224$) compared to the CIFAR-100 dataset (input resolution is $32\times 32$). However, the number of parameters remains the same across different datasets since it depends only on the model's architecture.

%\section{Code}
%Code and pre-trained models have been uploaded in the zip files.

\end{document}